# FREIDA: A Framework for developing quantitative agent based models based on qualitative expert knowledge

Frederike Oetker[1*], Vittorio Nespeca[1], Rick Quax[1,2]

# Abstract

Agent Based Models (ABMs) often deal with systems where there is a lack of quantitative data or where quantitative data alone may be insufficient to fully capture the complexities of real-world systems. Expert knowledge and qualitative insights, such as those obtained through interviews, ethnographic research, historical accounts, or participatory workshops, are critical in constructing realistic behavioral rules, interactions, and decision-making processes within these models. However, there is a scarcity of systematic approaches that are able to incorporate both qualitative and quantitative data across the entire modeling cycle. To address this, we propose FREIDA (FRamework for Expert-Informed Data-driven Agent-based models), a systematic mixed-methods framework to develop, train, and validate ABMs, particularly in data-sparse contexts. The main technical innovation introduced within this framework is the extraction of what we call Expected System Behaviors (ESBs) from qualitative data, which are testable statements evaluated through model simulations. Divided into Calibration Statements (CS) for model calibration and Validation Statements (VS) for model validation, ESBs underpin a rigorous evaluation mechanism on the same footing as quantitative data. By structuring qualitative insights as explicit model constraints, FREIDA creates a transparent foundation for applying established modelling practices such as Sensitivity Analysis (SA) and Uncertainty Quantification (UQ), allowing modellers to assess parameter influence, model robustness, and remaining uncertainties in a systematic manner. Through this, qualitative insights can inform not only model specification but also parameterization, validation, and continuous improvement of model reliability and fitness for purpose, addressing a long-standing challenge in agent-based modeling. We illustrate the application of FREIDA through a case study of criminal cocaine networks in the Netherlands.

[1]Computational Science Lab, Institute of Informatics, University of Amsterdam, Amsterdam, Netherlands
[2] Institute for Advanced Study, University of Amsterdam, Amsterdam, Netherlands

*Corresponding author Frederike Oetker, f.oetker@uva.nl

# Introduction

Agent-based modeling has emerged as a powerful approach for understanding complex systems across various disciplines, from social sciences to ecology and economics. However, many research domains struggle to take full advantage of computational modeling approaches due to challenges in quantitative data availability and quality. The traditional modeling cycle relies heavily on comprehensive quantitative datasets, however many real-world applications, such as healthcare decision-making, community resilience, or social support networks, are characterized by the possibility to collect rich qualitative insights while presenting limited quantitative measurements. This fundamental challenge has historically restricted the application of agent-based models in domains where human behavior, social dynamics, and contextual factors play crucial roles (Aamodt and Plaza 1994; Voinov and Bousquet 2010).

The integration of qualitative data and expert knowledge into agent-based models, however, presents both opportunities and methodological challenges. While qualitative research provides rich contextual insights into agent behaviors, decision-making heuristics, and multi-actor interactions, it simultaneously introduces operational obstacles. Specifically, it is challenging to translate qualitative insights that are nuance-rich, descriptive, and subject to distinct subjective interpretations into the objective, unambiguous algorithmic code and exact parametric boundaries required by a computational simulation. This translation risk often results in a "loss of data fidelity," where the systemic constraints or cognitive rules intended by domain experts are distorted or oversimplified during hardcoding. Despite growing interest in integrating qualitative evidence into agent-based modeling, existing approaches primarily use qualitative information during model conceptualization and lack systematic methods for incorporating such evidence into the quantitative calibration and validation stages of model development. As a result, qualitative insights from interviews, expert narratives, and case files often remain interpretive inputs rather than operational modeling constraints, limiting their ability to systematically inform parameter estimation, uncertainty analysis, and model evaluation. A number of methodologies have been introduced to address this challenges such as Bharwani et al.'s Knowledge Elicitation Tools (KnETs) for deriving agent rules through mixed methods, Neumann et al.'s hermeneutic modeling qualitative approach (Neumann, 2023), Ghorbani et al.'s structured interpretation of qualitative data through the MAIA meta-model (Ghorbani et al., 2013; 2015), and Nespeca et al.'s combined use of two conceptual frameworks (one focused on the specific category of policy problems and another centred on agent behaviour through generic models) to balance cross-case comparability and the search for generalizable insights with the flexibility needed to capture the nuances of individual cases. (Nespeca, Vittorio, Comes, Tina and Brazier, Frances 2023). These methodologies demonstrate how qualitative methods (sometimes in combination with quantitative techniques) can be integrated within systematic workflows for developing agent-based models in complex policy environments where quantitative data alone may be insufficient.

As (McElreath and Smaldino 2015) argue, the integration of qualitative data e.g., capturing expert knowledge can occur at multiple phases of model development, evaluation, and use cycle, such as when informing model conceptualization, specifying behavioral rules, parameterizing interactions, or validating outcomes. An effective modeling approach that can be implemented in practice is one that covers and integrates all such different phases into a systematic methodology. However, although various methods facilitate the integration of qualitative data into computational agent-based modeling for some of these levels, two main gaps remain: (i) the explicit inclusion of qualitative knowledge within standard, mathematically driven quantitative model calibration and validation procedures, which shifts qualitative findings from being merely interpretive inputs into systematically-applied, operational modeling constraints which provides the structured parameterization and documentation required for applying established Uncertainty Quantification (UQ) and Sensitivity Analysis (SA) approaches to investigate how variations in qualitatively derived parameters influence model behaviour and reliability; and (ii) the synthesis of a modeling pipeline from conception to computational model that covers all the levels while systematically embedding these UQ and SA protocols throughout the entire development lifecycle.

Our proposed FREIDA framework addresses these gaps by introducing an end-to-end mixed-methods pipeline that transforms qualitative evidence into systematically applied, operational modeling constraints rather than leaving them as merely interpretive inputs. The framework's core methodological innovation lies in Expected System Behaviors (ESBs), which systematically translate unstructured findings from interviews, case files, and expert narratives into formalized Calibration Statements (CSs) and Validation Statements (VSs) to directly guide parameter estimation, rigorous UQ, and model evaluation. Validation Requirements (VRs) and Thresholds assist in guiding the model evaluation. Unlike existing approaches that primarily restrict qualitative information to initial model conceptualization, the ESBs approach leverages Thematic Content Analysis (TCA) to construct testable structural statements that evaluate and score model executions. By integrating these qualitative behavioral boundaries with standard empirical data-fitting metrics (e.g., mean-squared errors) into an aggregate scoring procedure and embedding systematic SA throughout the development lifecycle, FREIDA makes rigorous computational modeling accessible to complex, data-limited domains historically considered ill-suited for empirical agent-based modeling approaches.

# Existing frameworks and research gap

A growing body of research has explored how qualitative evidence can support the development of Agent-Based Models (ABMs), resulting in a diverse methodological landscape encompassing knowledge elicitation tools, conceptual modelling frameworks, mixed-methods approaches, and validation strategies. However, these approaches tend to support specific phases of the modelling lifecycle rather than providing a fully integrated methodology spanning conceptualization, calibration, validation, and uncertainty analysis.

Several methodologies focus primarily on knowledge elicitation and conceptual model development. Bharwani et al.'s Knowledge Elicitation Tools (KnETs) support the extraction of behavioural rules from qualitative evidence and their translation into formal model specifications (Bharwani et al. 2015). Neumann's hermeneutic modelling approach emphasizes narrative and interpretive methods to capture cultural meaning and enrich agent representations (Neumann 2023). Complementing this perspective, Nallur et al. (Nallur, Aghaei, and Finlay 2024) argue that elicitation strategies should be adapted to different stakeholder groups to improve the integration of expert knowledge into ABM development.

Other frameworks focus on structuring and formalizing qualitative information. Ghorbani et al.'s MAIA framework provides a systematic approach for organizing institutional, behavioural, and interactional data into formal model structures (Ghorbani et al. 2013). Importantly, subsequent work by Ghorbani, Dijkema, and Schrauwen (Ghorbani, Dijkema, and Schrauwen 2015) also addresses conceptual and empirical validation, emphasizing the importance of ensuring consistency between model assumptions, qualitative evidence, and observed system dynamics. Likewise, Epstein's work on constructing quantitative models from qualitative narratives demonstrates how qualitative observations can be translated into formal computational representations, thereby providing an important foundation for integrating narrative evidence into ABMs (Dixon, D.S. and Reynolds, W.N. 2005; Epstein 1999).

Research has also explored the role of qualitative evidence in calibration and validation. Castellani et al. (Castellani, Barbrook-Johnson, and Schimpf 2019) employed a mixed-methods validation approach combining simulation outputs, empirical observations, and expert feedback. Seidl (Seidl 2014) discusses how qualitative data may inform multiple stages of model development and validation, while Yang and Gilbert (Yang and Gilbert 2008) advocate moving beyond purely numerical data sources by incorporating qualitative observations into agent-based modelling. Wijermans et al. (Wijermans et al. 2022) further examine the integration of multiple forms of evidence, particularly behavioural experiments, into ABM development and evaluation, with Antosz et al. (Antosz et al. 2022) providing an overview of using agent-based simulation for this purpose.

Recent methodological advances have additionally explored how systems-thinking approaches can support quantitative modelling workflows. For example, Crielaard et al. (Crielaard et al. 2022) introduced annotated Causal Loop Diagrams (CLDs) and Multi-Model Structures to improve expert communication, identify mediating mechanisms, and facilitate the formalization of qualitative system relationships into quantitative representations. While originating from System Dynamics rather than ABM methodology directly, these approaches demonstrate how qualitative system knowledge can support formal model development.

Finally, some studies have focused on operational and policy-oriented validation. Manzi and Calderoni's MADTOR model (Manzi and Calderoni 2024), designed to simulate the adaptation of drug trafficking organizations to law enforcement interventions, demonstrates the importance of operational validation by comparing simulation outputs with empirical behavioural patterns. The framework also illustrates the value of scenario-based experimentation for evaluating intervention strategies. Even so, MADTOR relies primarily on quantitative evidence and does not provide a systematic framework for integrating qualitative information throughout calibration and validation.

Taken together, these approaches demonstrate the growing recognition that qualitative evidence can contribute meaningfully to multiple stages of ABM development. However, the current methodological landscape remains fragmented. Existing approaches typically address isolated phases of the modelling lifecycle, such as conceptualization, rule elicitation, or validation, without providing a unified methodology that systematically integrates qualitative evidence throughout the entire modelling process. In particular, there remains a lack of end-to-end frameworks that combine qualitative knowledge elicitation with formal calibration procedures, operational validation, and uncertainty and SA. This gap is particularly important in domains characterised by sparse, uncertain, or partially observable data, such as organized crime and illicit networks, where qualitative evidence often constitutes a primary source of system knowledge.

# Proposed Framework

The proposed framework consists of three phases: Knowledge and Data acquisition, Integration of data on the conceptual and computational model, and Validation of the model. These phases are presented in Figure 1 and summarized in the following.

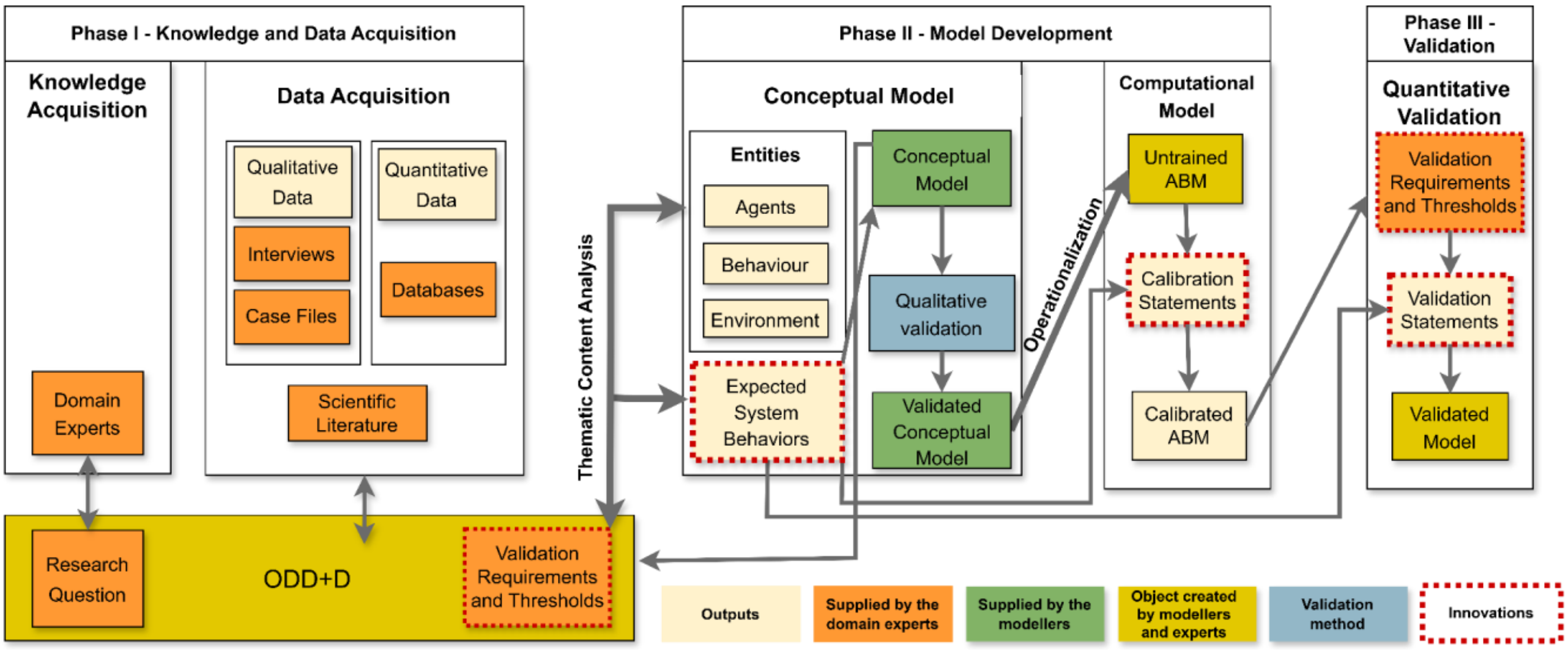

**Figure 1:** We outline the FREIDA framework, and its three phases for developing and validating an agent-based model (ABM). Phase I involves defining the research question and gathering qualitative and quantitative data through domain experts, literature, and case sources. Phase II focuses on building a conceptual model (agents, behaviors, environment), validating it qualitatively, and then creating and calibrating a computational model. Phase III applies quantitative evaluation using predefined thresholds to assess whether the model meets specified validation requirements. The outcome informs whether further model refinement is required. Color coding indicates contributions from domain experts (orange), modellers (green), shared outputs (yellow), validation methods (blue), and key innovations (red dashed outlines). Noteworthy, VRs and Thresholds are present both in Phase I and Phase III, as they are created in the former, and applied in the latter.

# Phase I: Knowledge and Data Acquisition

The goal of this phase is the collection of the knowledge, data, and domain experts pool necessary to inform the model development.

The 'Knowledge Acquisition' step is an iterative loop between identifying the relevant domains of expertise, corresponding domain experts pool, and the model's purpose. The model's purpose is defined by specifying the research question it aims to address and outlining the context. Context refers to the system and time boundaries within which the model is to be relevant and valid, aligning with the experimental frame (Railsback and Grimm 2019). System boundaries determine the model's scope by specifying which processes, entities, and interactions are included or excluded, while time boundaries define the time step and the simulation period. In addition to defining the model's purpose and scope, we propose that the Purpose section of the ODD+D protocol include initial VR. These early requirements should articulate high-level expectations for model performance, which can later be formalized into quantitative validation thresholds. This iterative process is typically initiated through preliminary discussions or think tanks with key stakeholders or initial domain experts, where the broad problem space is explored and refined, often drawing upon established participatory modeling techniques to progressively define the model's purpose and identify the necessary expertise pool (Grimm et al. 2014; Müller et al. 2013). We recommend that this process is facilitated by a structured process, such as participatory workshops or Delphi methods (Gray et al. 2018; Norström et al. 2020), (Franco and Montibeller 2010) utilizing the ODD+D framework (Müller et al. 2013) to guide the process as well as keep track of the progress. Our second recommendation is the utilization of what we call an expertise table, which is essentially a contingency table listing the currently involved domain experts (rows) and maps them to relevant expertises (columns) to ensure that sufficient coverage is achieved (Uleman et al. 2021). The ODD+D approach combined with an expertise table enables systematic iteration until convergence on team composition and ODD+D document.

Following this, in the 'Data Acquisition' step, data sources (both quantitative and qualitative) are identified.

## Information identification

The initial phase of the FREIDA framework focuses on the crucial collection of knowledge, data, and the identification of a domain expert pool necessary to inform the model development. This phase establishes the model's purpose, often expressed through a research question and a clearly defined context of application (e.g., criminal networks in Amsterdam). The modeling context includes both system and time boundaries, collaboratively defined during structured sessions with a panel of domain experts. System boundaries determine which agents, interactions, and processes are considered relevant, while time boundaries specify the temporal resolution and simulation period. These elements guide data collection, agent design, and the intended scope of generalization.

To initiate this process, we conduct expert focus groups based on a structured protocol (see Table 4 in Appendix I). The goal is to iteratively refine the problem space, determine which expert domains are relevant, and establish the foundational ODD+D document (Müller et al. 2013). This protocol enables mutual learning between domain experts and modelers, drawing on participatory modeling principles and structured elicitation tools (e.g., the expertise table, which maps expertise domains to individuals and helps identify knowledge gaps) (Uleman et al. 2021). ODD+D serves as the overarching documentation structure to facilitate and track progress.

Once the modeling context is defined, the next crucial step involves collecting the data necessary to inform the development of a preliminary ODD+D document, alongside any additional data sources required for subsequent steps of the FREIDA framework. This includes both qualitative data (e.g., interview transcripts, ethnographies, police reports) and quantitative data (e.g., demographic statistics, network snapshots). Interviews with experts, both from the focus groups and via snowball sampling, form the backbone of the qualitative data collection. These interviews follow a simplified ODD+D structure to ensure comprehensiveness while remaining accessible to non-technical participants (see Table 1 in Appendix I for the protocol.

Importantly, data should not be collected in isolation from the modeling goals. It is essential that the data fits the model's context and collectively provides adequate coverage of the identified domains. Where external datasets are considered, their relevance must be explicitly justified by expert contextualization. This ensures that any auxiliary data remains theoretically meaningful and coherent with the defined system boundaries. We recommend continuing data collection until theoretical saturation is reached, and no new relevant insights are emerging (Scott and Glaser 1971; Guest, Bunce, and Johnson 2006). At this point, the combined insights from expert input and available datasets allow modelers to produce an initial version of the ODD+D document, which outlines agents, behaviors, environmental components, and high-level processes.

## Thematic Content Analysis

Between Phase I and Phase II, we initiate a transitional process that integrates the structured ODD+D framework (Müller et al., 2013) with Thematic Content Analysis (TCA) (Braun and Clarke, 2006; Boyatzis, 1998). After defining the research question, identifying data sources, and engaging domain experts, the next challenge is systematically extracting key concepts such as agents, behaviors, and structures from qualitative data and determining how this can be achieved with rigor. This is where TCA becomes essential, as it enables us to derive structured, model-relevant insights from unstructured qualitative sources like interview transcripts, documents, and case files.

TCA is a widely used qualitative research method for identifying, analyzing, and reporting patterns (themes) within data (Braun and Clarke 2006; Boyatzis 1998). It is a flexible approach applicable to various data types, including interviews, focus groups, documents, and visual materials. Rather than merely counting words, TCA interprets underlying meanings and patterns (Naeem et al. 2023). The method enables researchers to transform raw narratives into structured codes, which are then synthesized into themes and used to generate meaningful insights for conceptual model development. The process involves data familiarization, initial coding, theme development, review, definition, and reporting.

In cases where ABMs are informed by complex, qualitative expert knowledge or heterogeneous data sources, a more structured approach is needed to ensure clarity, traceability, and reproducibility. This is why we employ ODD+2D as an additional option alongside ODD+D. While ODD+2D is designed to integrate both qualitative and quantitative data, its structured approach to data description and analysis provides valuable insights even when primarily qualitative data is available (Laatabi et al. 2018). While ODD+D supports the integration of qualitative domain knowledge into the core ODD framework, ODD+2D extends this further by offering dedicated components for handling both qualitative and quantitative data in a consistent and transparent way. This makes it particularly valuable when qualitative data is crucial in model design. The ODD+2D framework, with its emphasis on detailed process descriptions and data integration, complements TCA by offering a systematic way to organize and interpret qualitative findings. Specifically, ODD+2D's emphasis on describing 'Scheduling' and 'Stochasticity' helps in identifying temporal patterns and probabilistic behaviors within qualitative data, such as narratives of event sequences or descriptions of decision-making under uncertainty. Furthermore, ODD+2D's 'Initialization' and 'Input Data' categories provide a structured way to document the context and sources of qualitative data, ensuring transparency and reproducibility in the analysis process. This is particularly useful when dealing with diverse qualitative data sources like case files and expert interviews, where clear documentation of data provenance is crucial.

The TCA process begins with the development of a coding scheme based on the ODD+D document and informed by the structured categories of ODD+2D, even if quantitative data is limited. This coding scheme typically includes first-level codes such as "agents," "behavior," and "environment," with more specific subcategories beneath them. Please find the coding scheme in Table 6 in Appendix I. Specifically, when analyzing the focus group data, the modelers apply the ODD+D derived coding scheme to the transcribed discussions, systematically assigning codes to excerpts that correspond to agents, behaviors, environment, and system dynamics as outlined in Table 4 in the Appendix. During development of the coding scheme, the data is examined for recurring themes, relationships, and patterns, which may involve analyzing code frequencies, co-occurrence patterns, and identifying novel elements.

The coding scheme itself undergoes iterative refinement, being expanded or adjusted as new insights emerge from the continuous analysis of the focus group's qualitative data, ensuring a thorough and nuanced understanding of the system being modeled. ODD+2D's focus on data integration and process description helps to ensure that the qualitative insights derived from TCA are structured and comprehensive, facilitating the transition to model development.

This analysis identifies key themes, patterns, and, importantly, ESBs, which describe anticipated system-level dynamics (Grimm and Railsback, 2012). Notably, TCA also informs the refinement of the ODD+D document, particularly in its 'Theoretical and Empirical Background' section, by incorporating model invariants discovered during the analysis. This iterative feedback loop, where TCA outputs update the ODD+D, is a novel strength of our framework, as it ensures a dynamic and data-driven conceptualization of the model. The TCA process bridges expert insights with model development by organizing qualitative data into structured categories. In this context, its key outputs, agents (types, roles, attributes), behavioral patterns (decision rules and influences), and environmental factors (contextual conditions), map directly to core model components.

### Expected System Behaviors: A Novel Output from TCA

A key novelty of our framework is that we enrich the TCA process with a new type of output: ESBs. These describe expected, falsifiable emergent patterns of the system as different scales, as opposed to individual agents or interactions, and will be used for model calibration and validation. An ESB is most useful when it is most discriminative about the system behaviour, to do so the ESB specifies a spatial and temporal scale on which the behaviour should operate. It is important to note that ESBs are distinct from the traditional codes and patterns that lead to the behavioral rules to be implemented. Instead, they describe expected (partial) system states after a given time scale, and a set of conditions under which the pattern will emerge.

To illustrate the difference between ESBs and behavioral rules, consider the following example of a traditional pattern: "Agents with high violence potential are more likely to initiate conflicts with other agents." This pattern directly informs the implementation of agent-level behavior in the model. In contrast, as an example an ESB might state: "A single value network with high average violence potential will likely disintegrate into disconnected components, within three months after a kingpin liquidation." This ESB describes an expected outcome, emerging from the collective interactions of agents over time. It doesn't dictate specific agent behaviors, but rather provides a benchmark for assessing whether the implemented behavioral rules produce the anticipated macro/meso-level dynamics.

# Phase II: Model Development

## Splitting ESBs into Calibration and Validation Statements

Similar to how the classic modeling cycle (Cooke 2018)splits quantitative data into two parts to serve model calibration and model validation, respectively, we split the ESBs into two disjoint sets: Calibration Statements (CSs) and Validation Statements (VSs). Both types of statements should be defined as logical predicates. These should define conditions on the model output that should evaluate to 'true' or 'false'. It is important that these statements can be tested against model output, i.e., the (not yet developed) model should plausibly produce the relevant output and the concepts used in the statements should be included in the coding of agents, behaviours, etc. A statement can either be one that must always hold, never hold, or it can be conditional upon certain initial conditions.

To determine whether an ESB should be classified as a CS or a VS, we employ a Scale Separation Map (SSM) (Bhattacharya et al. 2021) which categorizes ESBs based on temporal (short-term vs. long-term) and spatial (localized vs. system-wide) scales. By positioning each ESB on this map, we systematically determine whether it should become a CS or a VS. That is, validation is typically performed to test generalizability of a model. Therefore, CS are derived from short-term and/or localized ESBs. This allows for the calibration to make sure that behavioral rules meet expectations by evaluating their (almost) immediate consequences. The added benefit is that running model calibration, which is a computationally expensive procedure, can be relatively efficient since no full-scale simulations are necessary to evaluate them. Conversely, VS are derived from long-term, system-wide ESBs, meaning that they describe emergent patterns at the macro level and are used to evaluate model validity. For example, the ESB "In case file A, when the original leader is removed, Agent Y assumes their role within 1 week." is a calibration statement, as it describes the result of interactions across agents at the meso level and in the short term. An example of a VS would be "After the liquidation of the kingpin, the entire network has fragmented into multiple smaller, disconnected components within 6 months.", since it describes long-term, system-wide effects.

Finally, although TCA provides the foundation for deriving VSs, these statements should be independently reviewed and approved by domain experts to ensure they reflect expected system behaviors rather than modeler bias. This step safeguards the objectivity of the expert-driven validation framework. Crucially, VSs must be distinct from calibration data; excessive overlap can inflate apparent model performance and undermine the credibility of the validation process.

## Conceptual model

Conceptual models define a model's key features qualitatively in order to guide its quantitative implementation. It is a process where modellers and domain experts work intensively together to give the best possible foundation for the quantitative model development, where the main effort shifts to the modellers.

The components of an ABM (Agents, Behaviour and Environment) can be broken down into entities and attributes, relationships (which can be linked to both Agents and Environment), rules and processes (comprising Behavior), and contextual variables (forming the Environment) (Jopp, Reuter, and Breckling 2011; Railsback and Grimm 2019). Entities are the key actors or components within the system, such as "customers" in a business model, "species" in an ecological model, or "actors" in a criminal network. Attributes describe the properties of each entity, which can be quantitative (e.g., age, population size) or qualitative (e.g., role type, species behavior). Relationships describe the structural connections between entities, which can be static (e.g., a criminal actor belongs to a specific market) or dynamic (e.g., a criminal actor changes role or social environment), while rules or processes dictate how these relationships evolve over time by defining the mechanisms that govern entity behaviors and interactions. In our framework, the definitions of these entities and attributes are obtained via the coding scheme during the TCA.

There are a number of existing methodologies that can support the formulation of the conceptual model, depending on the complexity and nature of the system being modeled. These approaches include amongst others the Belief-Desire-Intention (BDI) model (Rao and Georgeff 1997), the MAIA framework (Ghorbani et al. 2013), and OCOPOMO (Scherer et al. 2015), and provide structured ways to represent agents, their motivations, and system dynamics. Visualization tools such as CLDs, activity diagrams, or sequence diagrams (UML) can further help articulate the relationships and processes involved (Kulkarni et al. 2021). An overview of these methodologies and visual tools is provided in Table 7 (Appendix I), which offers practical entry points for readers seeking hands-on guidance beyond the high-level summary presented here. Regardless of which approach is selected, the resulting conceptual model should be sufficiently concrete to enable qualitative validation and eventual operationalization into a computational model. This includes clearly articulated assumptions, units of measurement, and narratives describing interactions and causal mechanisms.

**Validating the conceptual model: Structural validation**

While it is not yet computational and cannot generate predictive outputs, the structure of a conceptual model can be validated in collaboration with experts before proceeding to the implementation phase. Structural validation examines whether the agents, behaviors, and environment within the conceptual model function as intended and align with theoretical expectations (Qudrat-Ullah 2005; Bridget 2009; Andrew Collins 2024). It focuses on qualitative evaluation, relying on expert review, logical consistency checks, and scenario-based assessments rather than numerical simulations. There are concrete criteria for evaluating structural integrity in conceptual models, such as: clarity and definition (all variables and causal relationships must be explicitly defined, ensuring that the model avoids ambiguous or vague elements), causal justification (each link between variables must be logically justified or empirically supported, rather than relying on intuition or assumption), completeness (the model must include all necessary causes and mechanisms to capture the essential dynamics of the system, avoiding oversimplifications) and consistency and directionality (causal relationships should be correctly represented, ensuring that no cause-effect reversals or tautological loops distort the model's logic) (Burns and Musa 2001; Sargent 2013; Tesfatsion 2007).

A key aspect of structural validation is ensuring that the model's feedback mechanisms function as intended. In ABMs, mechanisms, such as threshold-based feedback rules, must be explicitly identified and tested to ensure that agents behave in a theoretically consistent manner. For instance, if an agent property is assumed to remain stable under certain conditions, but no stabilizing mechanism is embedded within the model, a structural mismatch arises between the conceptual design and its intended function. Such inconsistencies suggest that additional feedback rules may be necessary to align the model's behavior with its theoretical assumptions.

If structural validation reveals inconsistencies, the conceptual model should first be refined before proceeding to operationalization. This may involve revising agent interactions, causal links, or system rules to enhance coherence. Many existing methodologies offer a structured guide for reassessing the model's design. For example, Quadrat-Ullah describes their validation process explicitly as iterative, and that structural validity provides a rigorous standard for establishing confidence in a system dynamics model, even if the model performs well in behavior validity tests (Qudrat-Ullah 2005), while Sargent states that developing a valid simulation model is an iterative process, involving the creation and refinement of multiple versions before achieving a satisfactory outcome, emphasizing repeated conceptual model validation until the model meets the desired standards, with validation and verification portrayed as ongoing activities integrated throughout the model's entire life cycle (Sargent 2013).

## Computational Model

Once the conceptual model is developed, the next step is translating the concepts and narratives into an executable code (operationalization). After this, the model's parameters should be adjusted so that the simulations become a closer representation of reality or the desired outcome (calibration).

### Operationalization and Calibration

Operationalization represents the bridge between conceptual and computational models, transforming qualitative constructs into variables or constants and narratives into mathematical expressions. This process also involves choosing algorithms (e.g., under which conditions is a mechanism activated), data structures (e.g., are agent objects attached to other (environmental) objects or do they have attributes specifying their location), etc.

We refer to existing frameworks to describe detailed steps. At a high level, Babbie provides a comprehensive introduction to conceptualization and operationalization in social research, offering valuable context for this process in ABM (Babbie 2020). De Vaus also offers important insights into the relationship between research design and the translation of theoretical concepts into measurable variables (de Vaus 2020). Some specific modeling languages such as SysML (Huang, Ramamurthy, and McGinnis 2007) and Agent Unified Modeling Language (AUML) offer formal mapping techniques and structured design tools to convert conceptual models into executable code (Sha, Le, and Panchal 2011; Bauer, Müller, and Odell 2001). Further alternatives include GeneSim (Anon n.d.; Van Hoecke 2016), which enables UML-driven translation for discrete-event simulations, and xtUML, which supports platform-independent modeling with direct code compilation (Cignoni and Paci 2012; Kosar, Bohra, and Mernik 2016). These frameworks offer visual and formal specification methods that support structural clarity, with added strengths in implementation-readiness and executable precision.

We then employ traditional calibration techniques to identify precise numerical values to these parameters to align the model's outputs with observed data. Shadish, Cook, and Campbell discuss the logic of causal inference in experimental and quasi-experimental designs, which underpins the importance of rigorous calibration in ensuring model validity (Shadish, W. R., Cook, T. D., & Campbell, D. T. 2002). The only difference here is that we use the CSs during this process. By using the CSs during calibration we provide a novel way to use expert knowledge in establishing parameters. Note that the CS can be divided in the traditional way to perform cross-validation. This approach opens the door to a mixed-method calibration strategy that formally combines quantitative data with qualitative, theory-driven expectations encoded in CSs, enabling more robust and interpretable parameter inference.

**Calibration in a mixed-method manner**

The integration of CSs based on expected system behaviors into the calibration process of ABMs requires a shift from traditional single-metric approaches to multi-objective calibration frameworks (Akbarpour, Govers, and Dignum 2023). Conventional calibration methods typically rely on quantitative goodness-of-fit metrics such as mean squared error or likelihood scores. However, the inclusion of CSs necessitates the construction of composite objective functions that account for both numerical accuracy and alignment with qualitative, system-level behavioral patterns (McCulloch et al. 2022). In this framework, calibration is formalized as an optimization problem where the objective function combines standard statistical criteria with performance scores derived from CSs, often operationalized as the proportion of CSs that evaluate to true under a given parameter configuration (Akbarpour et al. 2023). This approach enables the systematic incorporation of expert knowledge regarding emergent behaviors while maintaining methodological rigor in parameter estimation (Smith, Rahmandad, and Eubank 2021).

The choice of calibration method is contingent upon the nature of the quantitative data and the structure of the CSs. In scenarios where data are abundant and CSs are well defined with clear temporal and spatial references, weighted composite objective functions offer an efficient and interpretable solution. Weights for different components of the objective function can be assigned through expert elicitation or optimized via cross-validation (Akbarpour et al. 2023). In contrast, Bayesian calibration techniques are well suited for applications involving parameter uncertainty, limited data availability, or complex, conditional CSs. These methods allow for the integration of heterogeneous sources of evidence within a coherent probabilistic framework (Jones, Gong, and Wang 2021). When CSs are highly non-linear or interact strongly with quantitative objectives, evolutionary algorithms and ensemble methods provide robust search capabilities across complex and multimodal parameter spaces. These methods do not require gradient information or restrictive assumptions about the underlying objective function (Moya, Chica, and Cordon 2021). For computationally intensive models, surrogate modeling techniques such as those based on random forests can be used to accelerate the calibration process while preserving fidelity (Lee, Zheng, and Bassetti 2024).

Successful application of CSs-integrated calibration frameworks depends on careful attention to scaling and normalization. Since quantitative fit metrics and CSs truth counts operate on different numerical scales, improper scaling can bias the optimization process and distort the calibration results (Smith et al. 2021). SA should be used to assess the influence of different weighting schemes on calibration outcomes. It is often advisable to use a two-phase calibration process, where initial efforts prioritize quantitative fit and subsequent refinements focus on improving CSs performance (Akbarpour et al. 2023; Jones et al. 2021). Moreover, a clear separation must be maintained between statements used for parameter estimation and those reserved for model validation. This distinction is essential to avoid overfitting and to ensure that the calibrated model retains predictive validity for out-of-sample or previously unobserved behaviors (Davis, Koehler, and Lynch 2024). By preserving this separation, the model can be robustly evaluated in terms of both its numerical fidelity and its capacity to reproduce emergent, system-level dynamics (Lee et al. 2024).

# Phase III: Validation

Quantitative validation performs simulations to evaluate VSs (true/false) as well as prediction accuracy of quantitative data not used in calibration (hold out data).

Often, quantitative validation ends here, where the ‘validation score’ should not deviate too much from the ‘calibration score’, for some arbitrarily chosen threshold. However, sometimes this alone is not sufficient, as it is not clear what fraction of ‘successful’ VSs (that evaluate to *true*) would constitute a model that is fit for purpose. This is especially true if the VSs and/or validation (hold-out) data are not chosen randomly but to assess a generalization power of the model, or are chosen to assess external requirements from the application domain, since in that case it is not even expected that a ‘valid’ model would achieve roughly the same score on the validation data as on the calibration data.

Indeed, validation in ABM development remains often an afterthought or is conducted in an ad hoc, non-precise manner (McCulloch et al. 2022). Despite the widespread adoption of standardised documentation protocols such as ODD and ODD+D, the specification of validation criteria frequently lacks the precision necessary for rigorous evaluation (An, Grimm, and Turner 2020). We argue that validation should be integrated into the model development process from the outset, with explicit, testable criteria that can be quantitatively assessed against simulation outputs. Otherwise, there is considerable risk that substantial effort goes into developing a model that, despite all good intentions, ends up never used in practice nor is deemed credible enough to be developed further by others.

Therefore, we propose enriching the Purpose section of the ODD+D protocol with specific VRs that establish clear expectations for model performance and utility. These requirements should be formulated as testable conditions that can be unambiguously evaluated as true or false during the validation phase, thereby creating a more transparent and rigorous assessment framework. We refer to these as validation thresholds, to make clear that the VRs must become testable. Note that these are meant to be overarching compared to the VSs: the VSs are evaluated against a single model execution, whereas the validation thresholds assess the statistics of the VSs across multiple model executions.

As an example, a VRs may at first be: “The model should be usable in practice for tactical decision making around which actor to follow/gather information on”. Translating this into a threshold, based on domain expert expertise of current practice, this could become: “The model should predict kingpin replacement correctly in at least 50% of our case files.”.

Recent literature highlights several validation approaches for ABMs. Collins et al. review nine such methods, ranging from foundational techniques like docking and empirical validation to advanced methods such as bootstrapping and causal analysis (Collins, Koehler, and Lynch 2024), stressing that validation should be purpose-driven, focusing on a model’s ability to reproduce meaningful patterns and processes relevant to the research question. Their proposed approach supports the development of VRs, which are explicit definitions of expected model behavior. These VRs can then be translated into Validation Tests (VTs) using quantitative metrics to assess whether the model meets its intended purpose.

Troost et al. introduce the "Keep It Adequate" (KIA) protocol, a twelve-step approach to ensure validity throughout the entire modeling process (Troost, Huber, et al. 2023). Rather than treating validation as a final step, they emphasize context-aware decisions from the beginning. The KIA protocol guides modelers in defining the modeling context, clarifying assumptions, and aligning validation criteria with the model’s argumentative purpose. Like Collins et al., they suggest that VRs should be grounded in the model's use-case and operationalized into concrete VTs to systematically assess adequacy.

Both approaches recognize that validity is contextual and purpose-dependent and offer complementary frameworks for formalizing validation. Collins et al.'s work presents a toolkit of methods that can be combined to create comprehensive VTs with specific thresholds, while Troost et al. offer a systematic process for ensuring adequacy throughout the modeling workflow (Collins et al. 2024; Troost, Huber, et al. 2023). It is even conceivable to develop a two-tiered validation framework, for instances where validation is of paramount importance: first defining abstract VRs based on the model's purpose and argumentative structure (following KIA), then translating these into concrete, testable validation thresholds using appropriate methods from Collins' toolkit. This would result in a set of clear true/false statements that could be systematically evaluated to determine if a model meets its VR (Troost, Huber, et al. 2023).

**Ensuring Generalizability and Model Robustness**

If model validation yields unsatisfactory results, several refinement strategies can improve model performance. Validation failures in agent-based models typically arise from two core issues: overfitting and insufficient generalization power. Overfitting occurs when a model performs well on calibration data but fails to reproduce meaningful patterns under new or unseen scenarios. This often results from excessive parameter tuning or overly complex model structures that capture noise instead of underlying system dynamics. Conversely, insufficient generalization arises when the calibration data lacks coverage of the diverse scenarios tested during validation. This situation is common when agents operate under fundamentally different conditions, social environments, or institutional constraints not represented during training (Collins et al. 2024). To diagnose overfitting, cross-validation techniques are recommended. Repeatedly partitioning the available data into calibration and validation subsets allows assessment of performance consistency. A small and stable difference between calibration and validation scores typically indicates good generalizability, while persistent discrepancies suggest overfitting. Monitoring learning curves during calibration offers further insight: if training error declines but validation error plateaus or increases, overfitting is likely (Troost, Berger, and Van Waveren 2023). This helps distinguish whether model complexity captures true structure or memorizes noise.

If overfitting is confirmed, dimensionality reduction through systematic SA is critical. Latin Hypercube Sampling (LHS) efficiently explores high-dimensional parameter spaces and uncovers interdependencies, overcoming limitations of simpler One-Factor-At-A-Time designs. Nearly orthogonal LHS configurations can sample parameter combinations comprehensively. Active nonlinear testing methods, including genetic algorithms, help identify critical parameter regions driving significant outcome variation. Metrics such as Partial Correlation Coefficients assist in ranking parameter influence and guiding dimensionality reduction efforts (Saltelli et al. 2008; Davis et al. 2024).

If overfitting is ruled out, poor validation likely reflects inadequate model specification or insufficient representativeness in the input data. This suggests missing key causal mechanisms or a calibration dataset too narrow in scope. In such cases, modelers should revisit earlier phases of the FREIDA framework, particularly Data Acquisition and Conceptual Model development, to improve system variability coverage. Incorporating new qualitative and quantitative insights at this stage enhances the model's ability to generalize and boosts predictive robustness in diverse contexts (Sargent 2013; Qudrat-Ullah 2005).

# Toy Model - Criminal Network

To address these challenges, structured approaches are needed that can translate qualitative expert knowledge into formal model components. The FREIDA framework is designed to support this process by providing a systematic pathway from expert statements to quantitative agent-based model calibration.

As an illustration, we have followed our own methodology to develop a minimal model in the domain of organized crime. It focuses on the scenario of replacing a (liquidated) 'kingpin' role in the network. This is a prime example where quantitative data offer little help, since kingpins are hardly recorded at all and even if they were, the replacement process is rarely captured by quantitative, observational data. We will describe key steps here insofar not already described in the general description above, and refer to the Appendix for output tables and other details where relevant. The selection process of relevant domain experts is skipped and two associated experts from the Police and intelligence domains were used as domain experts in the team.

## Training and validation

Our data is purely qualitative (for illustration) and consists of case files A-J, which can be found in Appendix I (with case files E-J used for calibration trajectories). Each describes a specific post-liquidation replacement scenario of up to one year, including key events and information such as roles of agents, (family) relationships, and other identifiers. The original case files cannot be made public but a synthesized case file is provided as an example (see Appendix I, Example Case File). Values and parameters such as trust, time until replacement, etc., are chosen arbitrarily by the modelers. They can be adapted as needed once the model is running, to produce results closer to real life networks.

**Boundary Conditions**

The boundary conditions are formed from key insights gained through filling in the ODD+D with the domain experts. They form the key steps of defining the problem statement. Boundary conditions for the toy model include for example the roles in the Value Network (VN) (described in the section Value Network in Appendix I), the size of the network or the parameters driving agents, such as trust.

*Network Size:* The problem statement focuses on mid-sized networks (20-50 agents).

*Agent Identity:* The model does not simulate ideological motivations, only profit-driven networks are represented by proxy, namely by assigning roles to agents and allowing them to try to connect to other agents with roles if their role pair is connected in the VN (see Figure 7 in the section Value Network in Appendix I).

*Tie Formation*: Ties are formed via necessity when agents seek to complete their individual Value Chains (VC), or through family of friendship bonds (trust). Both can change in reaction to a law enforcement intervention, such as removal of an agent. Reactions such as formation or cutting of a tie are triggered by such events.

*Law Enforcement Intensity:* Law enforcement interventions are out of scope. While the model captures kingpin removal, it does not account for sustained multi-target interventions.

*Temporal Scope:* Simulations reflect short- to medium-term network adaptation (up to 1 year post-intervention).

**Model development**

The toy model was developed using the boundary conditions derived from the ODD+D. Development followed a hybrid process of structured and unstructured expert elicitation, iterative conceptual modeling, and computational implementation.

Following the initial focus group sessions that produced the ODD+D framework, unstructured interviews with two domain experts were conducted to address gaps and ambiguities in the case files and preliminary conceptual model. These interviews targeted three main areas:

*Agent topology*: Refining the structural arrangement of agents, including typical degree distribution and clustering patterns.

*Role-specific behaviour*: Detailing how agents in different roles respond to interventions, pursue value chain completion, and manage trust relationships.

*Contextual conditions*: Clarifying the broader operational setting of the network, including common triggers for reorganization.

The ODD+D outputs define the model's purpose, entities, and processes in a formal structure. Key design choices included representing only profit-driven roles, and modeling tie formation through value chain necessity or trust. The process overview defined in the ODD+D (Appendix II) was the input for what became the agent interaction loop in the toy model, including the eight-step replacement sequence triggered by kingpin removal. TCA translated qualitative statements from the case files into concrete modeling rules, parameters, and CSs/VSs. First-level codes (Agents, Behaviours, Environment) and their second-level refinements ensured that the conceptual model was traceable to empirical knowledge. CS were operationalized as model invariants or short-term outcomes, while VS informed the target patterns for medium-term adaptation. From these combined sources, the conceptual model was implemented as a computational ABM. The implementation adopted minimal complexity to remain transparent and interpretable, while retaining the essential structural and behavioural elements identified by experts. Parameters such as trust thresholds, replacement time delays, and candidate selection rules were initially set to arbitrary but plausible values, then adjusted through expert review of early runs to improve face validity.

**Calibration and Validation Statements (CS and VS)**

ESBs were distilled from case files A-D and visualized in an SSM (Figure 2). The separation into CS and VS was decided based on the scales and relative quantities (2:1 split ratio). The complete list of CS and VS are detailed in the Appendix in Table 11.a. and b. and Table 12.a. and b., respectively; some examples are mentioned in the caption of Figure 2 while referring to their position on the SSM.

**Validation Requirements (VRs) and Thresholds**

The model's performance is validated by comparing simulation outcomes against empirical patterns from the case files (A-J). Since this is an illustrative toy model, we adopt a validation (fitness for purpose) threshold purely for demonstration purposes, if the model reproduces at least 85% of VS across multiple simulation runs. In practice, appropriate thresholds depend on the intended application and required confidence level, which is outside the scope of this paper. This amounts to reproducing core outcomes like leadership emergence, trust shifts, and network reorganization. Validation is measured across 48 simulation runs since the ABM is stochastic, and success is defined by the unweighted proportion of VS that the model satisfies.

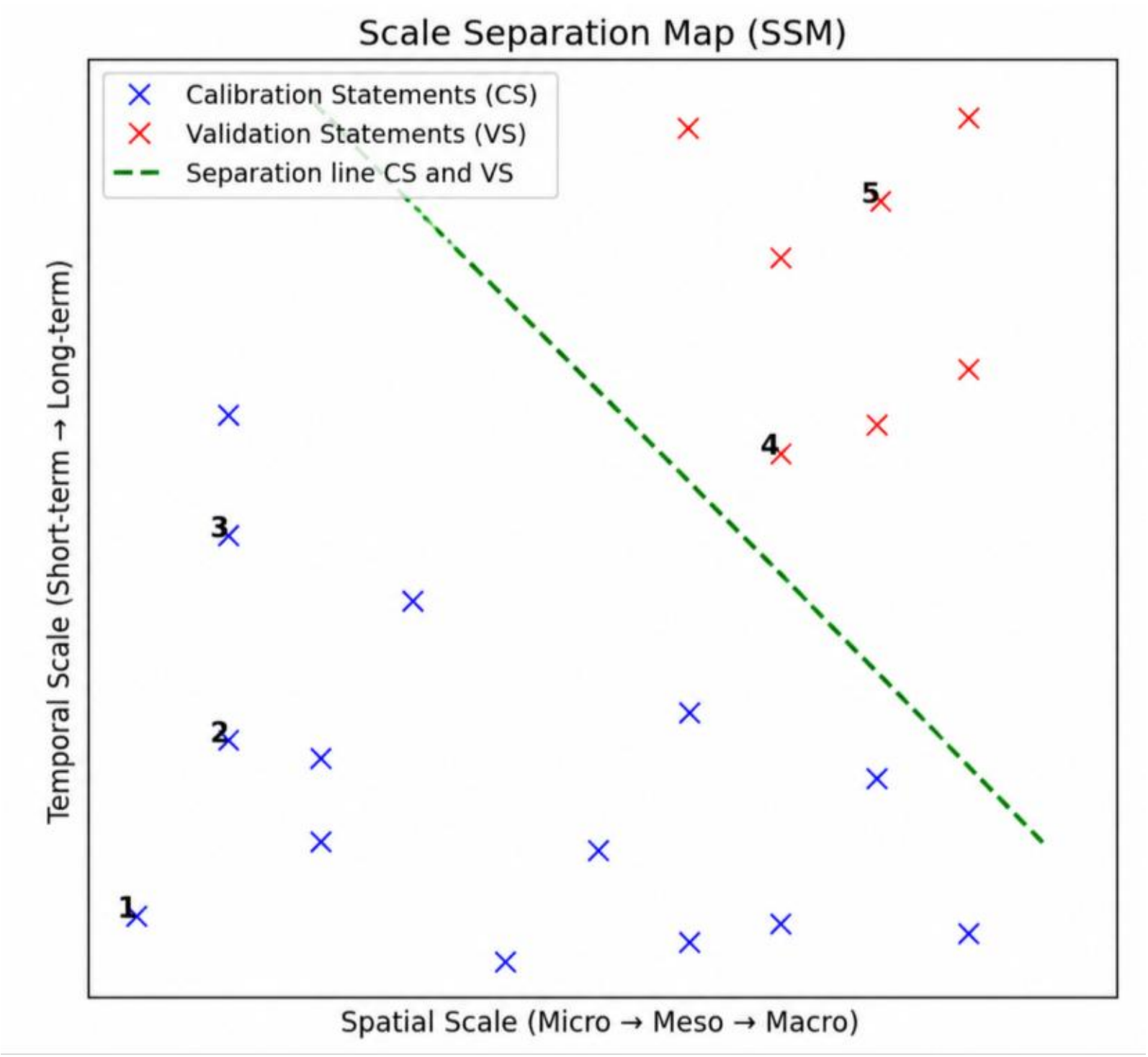

**Figure 2:** Scale Separation Map (SSM) illustrating the distinction between Calibration Statements (CSs) and Validation Statements (VSs). CSs (blue) represent short-term, localized agent behaviors used to calibrate the model, such as individual agent transitions or interactions. VSs (red) focus on long-term, emergent system behaviors, assessing the model's ability to capture global dynamics and generalize to real-world scenarios. The green dashed line separates CSs and VSs based on their respective spatial and temporal scales, with CSs addressing immediate behaviors and VS validating broader system patterns over time. Points 1-5 represent real CSs and VSs that can be found in Tables 11 and 12 respectively. Point 1: CS Case File A, VI: "The average violence capital among the orphans increases after the liquidation (measured at 1 week after)." This statement addresses the short-term behavior of the network after the removal of the leader, where the average violence capital of agents increases, which aligns with Agent Y taking a leadership role. Point 2: CS Case File C, Statement VI: "The trust between family members has an average of at least 0.75." This statement reflects how the local network structure adapts after a major event like liquidation. The short-term fluctuations in trust and connectivity indicate how smaller groups reorganize before a new leadership structure emerges. Point 3: CS Case File A, Statement IV: "A trusts B and C the most."This statement captures the trust dynamics in the network, which are essential for cooperation. The short-term nature of trust and cooperation is reflected in how Agent B and D would form a bond due to shared goals. Point 4: VS Case File D, Statement V: "The trust between the orphans increases by 15% before the new replacement is chosen" This VS reflects a long-term outcome of the liquidation event (the kingpin's removal), where individuals involved in the network's leadership are no longer present. Point 5: VS Case File D, Statement I: "Y is the new murderbroker by the end of the simulation." This VS validates the long-term emergence of a new leader (a new kingpin) following the removal of the previous one, reflecting shifts in group dynamics over an extended period.

The entities, ESBs and VRs then make up the conceptual model.

## Model calibration based on CSs

CS guide parameter adjustments to align the model with real-world observations. Each statement represents a quantifiable characteristic or event that the model must reproduce. Accurate prediction earns one point toward the total calibration score. Parameter values are iteratively updated until no further improvement in calibration score is observed for an extended number of steps. The result is a calibrated model whose parameter values provide the best fit to the available qualitative evidence under the chosen calibration procedure.

Only case files A-D were used to distill EBSs, each case file contributing a defined number of calibration points, which is meant to ensure that the model does not 'overfit' on a specific scenario. Although four case files are insufficient to fully prevent overfitting, they are adequate for the illustrative application. A second way in which overfitting can be countered is to further split the CS into smaller chunks (cross validation).

### Calibration identifiability and parameter uncertainty

To assess how empirical qualitative evidence constrains model parameters within the FREIDA framework, we examine the variability of calibrated parameter values across repeated optimization runs (Figure 3). Although all parameters are theoretically informed by expert knowledge, the results demonstrate that only a subset consistently converges to narrow ranges. Other parameters exhibit substantial variability without clear optima. This distinction highlights an important feature of expert-informed ABMs: while qualitative knowledge can meaningfully structure model dynamics, not all parameters are equally identifiable from available evidence. FREIDA explicitly accommodates this by encouraging transparency regarding which parameters are robustly informed and which remain weakly constrained.

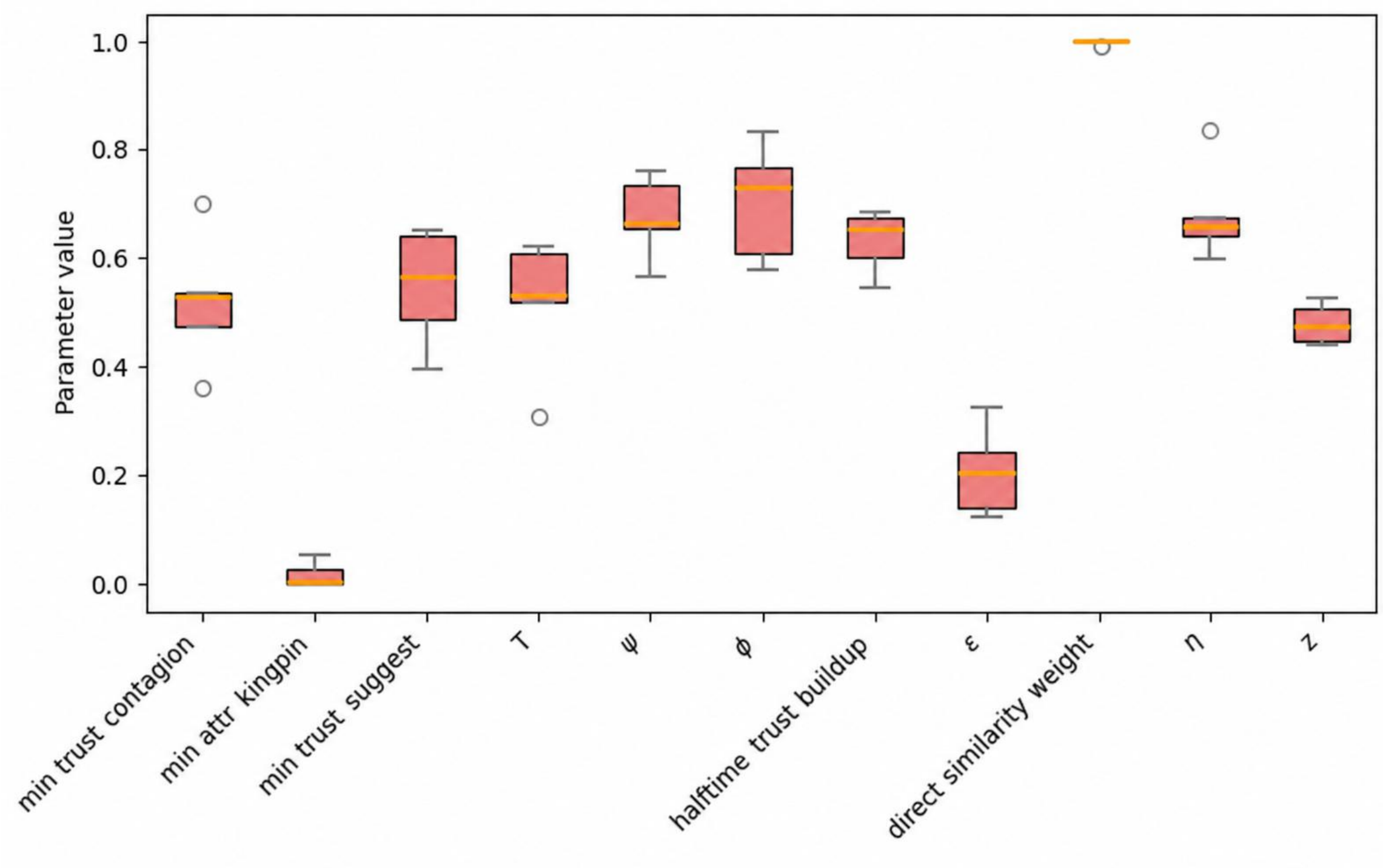

**Figure 3:** Distribution of normalised parameter values obtained from repeated calibration runs of the toy model within the FREIDA framework. Boxplots show the variability of calibrated parameters across runs after scaling to the unit interval. Narrow distributions indicate well-constrained parameters, while wider distributions reflect parameter uncertainty and limited identifiability from the available empirical evidence. Please refer to Table 10 and Equation 1 for a comprehensive overview of the parameters in the toy model.

### Heterogeneity across qualitative case files

FREIDA emphasizes that expert knowledge is often heterogeneous, context-dependent, and uneven in its informational content. This is reflected in Figure 4.a., which shows calibration trajectories obtained when optimizing the model separately on individual qualitative case files. Calibration outcomes vary markedly across cases, both in terms of convergence speed and final loss. These differences indicate that case files encode distinct and sometimes conflicting signals about the underlying processes. Rather than treating this heterogeneity as noise, FREIDA treats it as a substantive property of expert data that must be explicitly examined.

### Aggregation versus disaggregation of expert evidence

A central design choice within FREIDA concerns whether expert-derived evidence should be pooled prior to calibration or treated disaggregately. Figure 4.b. compares these approaches by contrasting calibration on aggregated case files with the average outcome of individual calibrations. While pooled calibration produces smoother loss trajectories and increased numerical stability, it does not necessarily yield superior fit. This illustrates a key trade-off: aggregation can dampen idiosyncratic biases but may also obscure structurally meaningful variation present in individual expert accounts.

**Internal consistency of qualitative knowledge**

Even within a single case file, expert knowledge may not be internally homogeneous. Figure 5 demonstrates this by calibrating the model on individual statements extracted from one case file. The resulting loss trajectories differ substantially, revealing that different statements emphasize different mechanisms or outcomes. This finding reinforces the importance of FREIDA's structured knowledge elicitation and encoding steps, which aim to reconcile such internal inconsistencies rather than implicitly averaging them away.

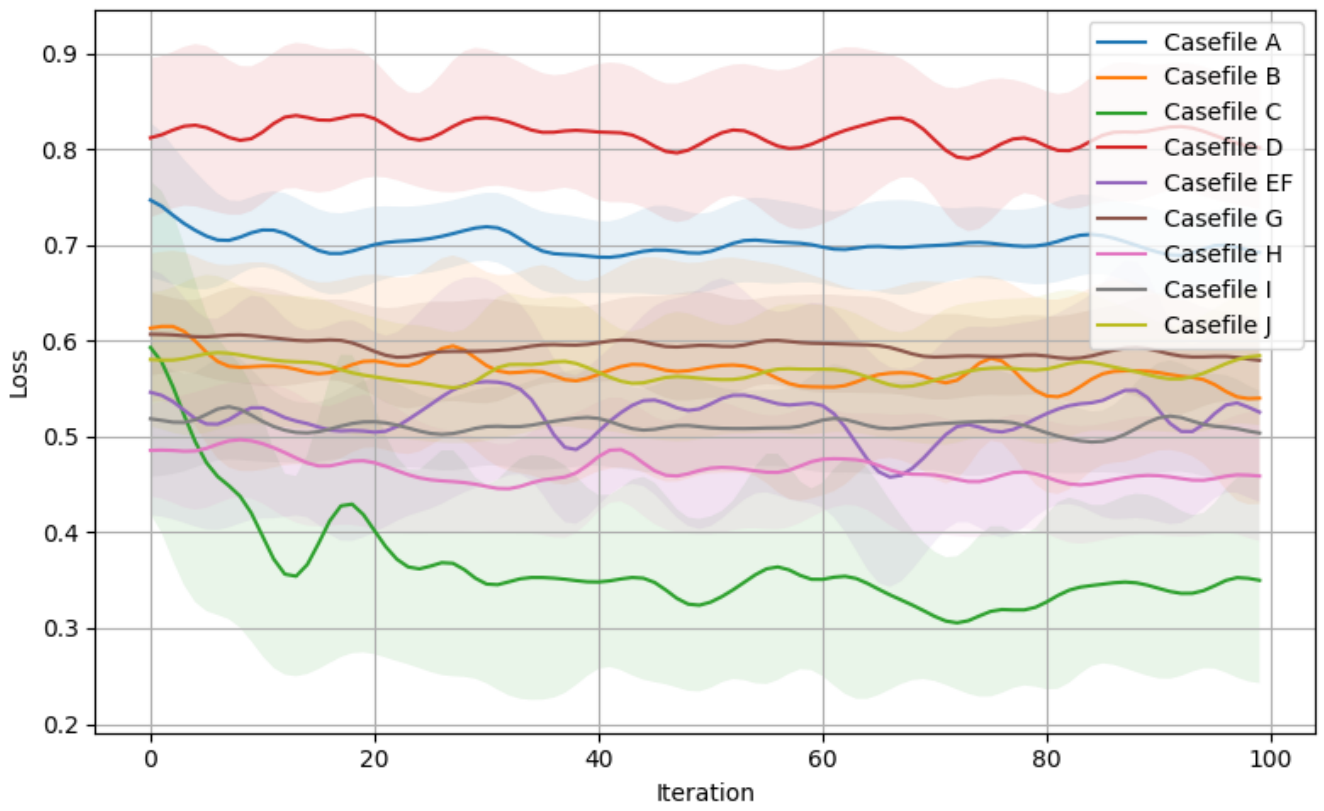

**Figure 4.a.**

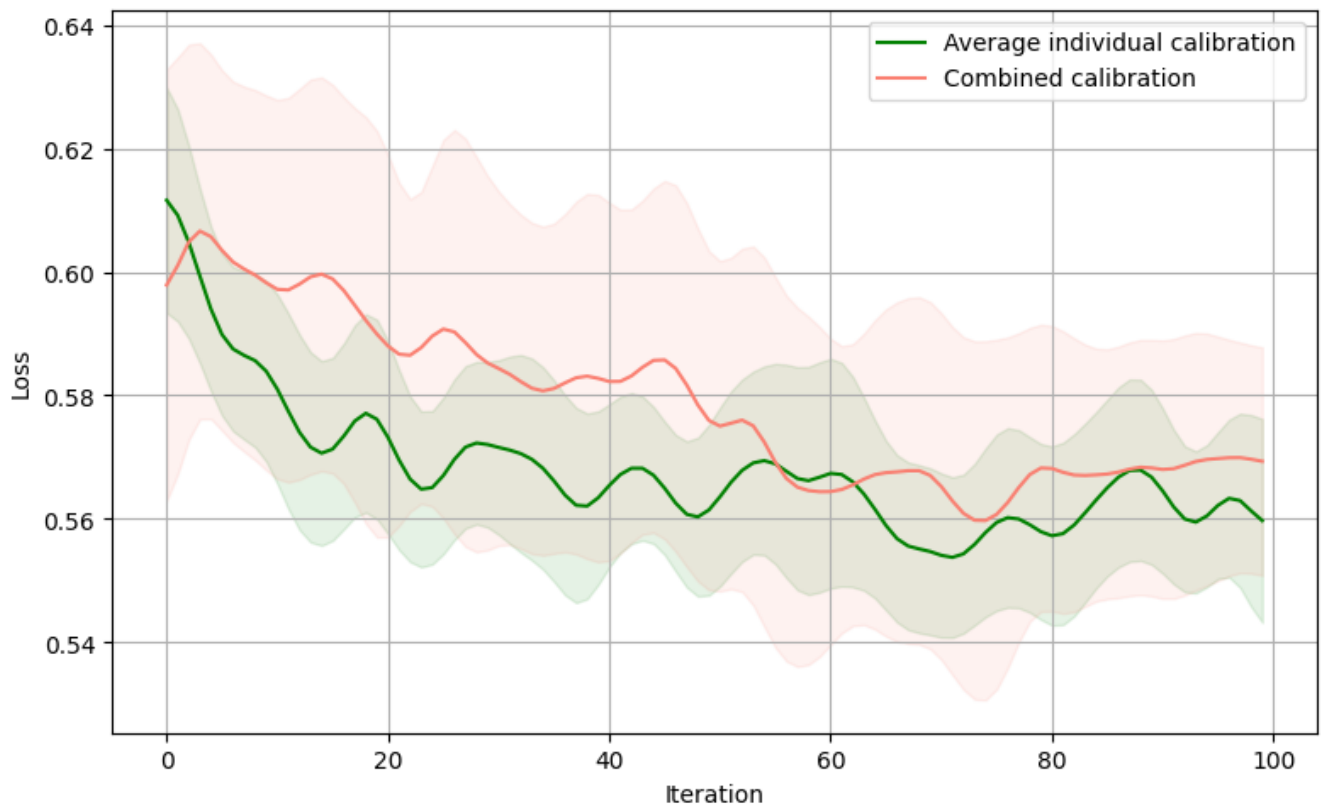

**Figure 4.b.**

**Figure 4.a.:** Calibration trajectories obtained when calibrating the toy model separately on individual qualitative case files. Each trajectory represents the evolution of the objective function (loss) averaged across multiple stochastic simulations, with shaded regions indicating the corresponding standard deviation. The figure illustrates substantial heterogeneity in calibration behavior and final loss values, demonstrating that calibration outcomes are strongly dependent on the specific expert-derived case file used. This variability highlights the sensitivity of the optimization process to differences in qualitative evidence across cases. For the exact statements per case file, please refer to Tables 11.a. and 11.b in the Appendix.

**Figure 4.b.:** Comparison between calibration using aggregated qualitative evidence and the mean loss obtained from calibrations performed on individual qualitative case files. The figure shows that calibration on pooled evidence yields more stable optimization trajectories, as reflected by reduced variance, but does not consistently achieve lower loss values than individual-case calibrations. This result illustrates the trade-off between robustness and case-specific fidelity when integrating heterogeneous expert knowledge into the calibration process.

In Figure 4.a., the calibration trajectories are presented as obtained when the toy model is calibrated separately on individual qualitative case files. The calibration statements of case files A-J can be found in Tables 11.a. and 11.b. in Appendix I. The non-flat (sloped) segments of the trajectories reflect periods in which the optimisation algorithm successfully identifies parameter updates that improve agreement with the calibration statements, resulting in decreasing loss values. Differences in slope across trajectories indicate that some case files provide stronger behavioural constraints than others, allowing faster or more pronounced improvements. In contrast, flat segments emerge when the optimisation process reaches plateaus or local optima where further parameter adjustments yield little improvement in loss. Another reason these plateaus arise can be that qualitative evidence may weakly constrain certain mechanisms, meaning that multiple parameter configurations perform similarly well. Consequently, the alternation between flat and non-flat regions illustrates both the learning dynamics of the optimiser and the heterogeneous informational content of the individual case files.

In Figure 4.b. we compare calibration based on aggregated qualitative evidence with the average loss trajectory obtained from individual-case calibrations. The aggregated trajectories appear smoother and exhibit reduced variability, reflecting the stabilising effect of pooling evidence across case files. Improvements in loss tend to occur more gradually, illustrating the trade-off between robustness and case-specific fidelity. Non-flat portions still occur when the optimiser identifies parameter changes that improve the overall fit to the combined evidence, but these improvements tend to be more gradual because conflicting signals across cases dampen strong directional updates. Flat regions here therefore primarily indicate convergence toward parameter areas that balance competing case-specific requirements, as well as reduced sensitivity of the aggregated loss to small parameter changes. Together, the presence of smoother slopes and extended plateaus highlights the trade-off introduced by aggregation: increased robustness and reduced variance in the optimisation process, but potentially weaker responsiveness to improvements that are specific to individual cases.

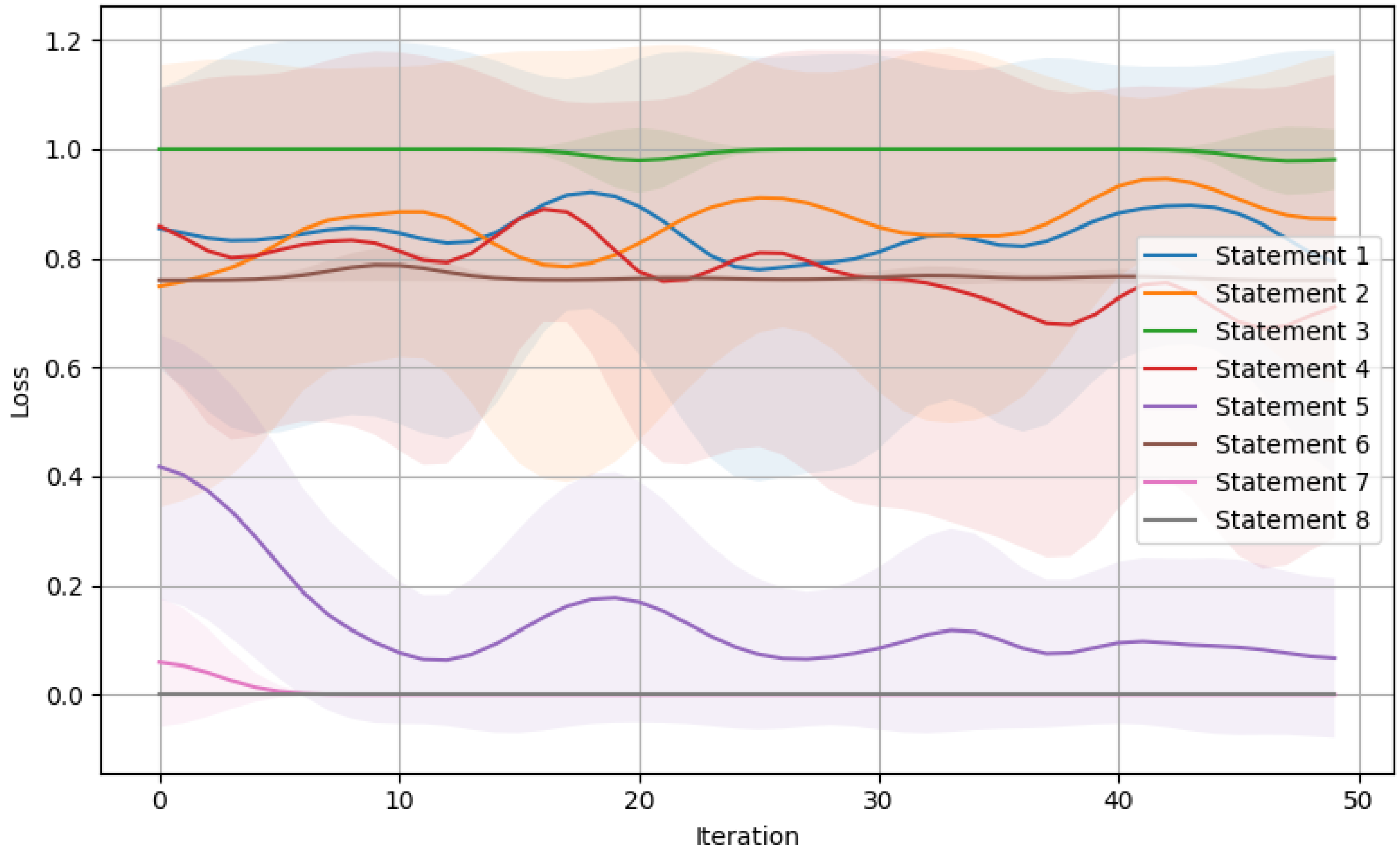

**Figure 5:** Calibration trajectories based on individual statements extracted from a single qualitative case file (case file B, the case of an intervention involving the liquidation of a high-profile cocaine trafficker). Results reveal significant variation in loss landscapes across statements, indicating internal heterogeneity within expert narratives and emphasizing the importance of structured aggregation within the FREIDA framework.

Figure 5 examines the internal consistency of qualitative evidence within a single case file by showing calibration trajectories derived from individual statements in case file B, concerning the liquidation of a high-profile cocaine trafficker. The trajectories vary in convergence speed and final loss values, indicating that statements constrain model behaviour differently. Extended plateaus suggest weaker or more ambiguous constraints, while steeper declines indicate stronger restrictions on parameter space. These variations demonstrate that expert narratives may contain heterogeneous or competing perspectives, highlighting the importance of FREIDA's structured elicitation, encoding, and aggregation procedures for explicitly reconciling uncertainty rather than averaging it away.

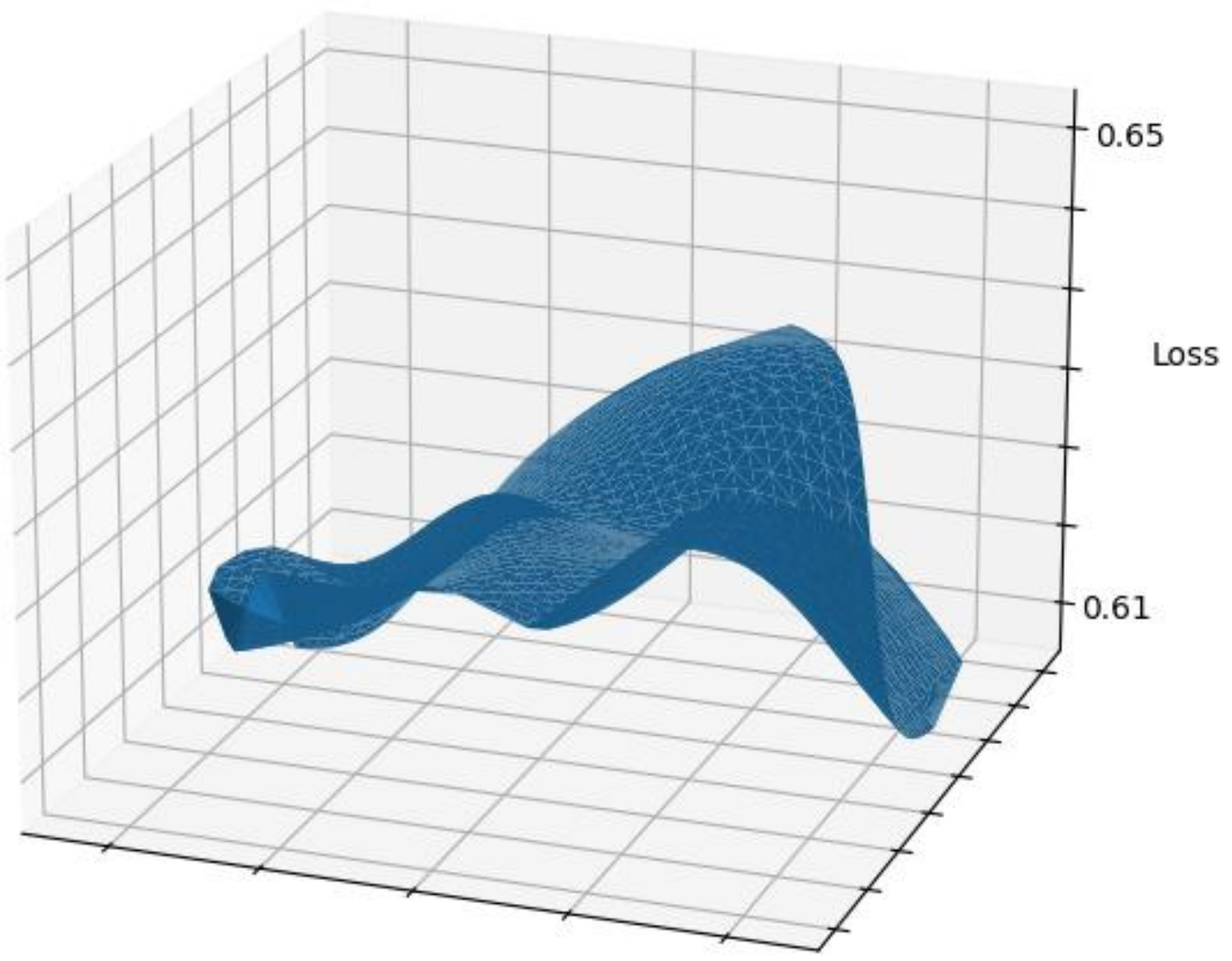

**Figure 6:** Objective function around the optimal solution. Robust, low-loss parameter regions with limited sensitivity to small changes are indicated, while mild gradients highlight calibration tensions that could guide further refinement.

Figure 6 provides a refined view of the objective function in the vicinity of the optimal parameter region. The landscape forms a shallow basin with loss values ranging between approximately 0.61 and 0.65, indicating limited sensitivity to small parameter perturbations around the optimum. The minimum remains located close to the centre of the explored slice, while neighbouring configurations increase the loss by roughly 0.02–0.04, equivalent to a modest increase in failed calibration statements across the case set. This smoother structure indicates that the toy model successfully constrains the parameter space to a coherent low-loss region while preserving uncertainty. Substantively, this implies that the ESBs-derived calibration statements meaningfully restrict model behaviour in this toy model implementation, yet do not enforce a single deterministic solution. The results therefore support an ensemble-based interpretation of calibrated models and highlight the importance of SA when qualitative evidence underdetermines precise parameter values.

In addition to revealing the qualitative structure of the optimum, the numerical scale of the basin indicates that deviations within the central region correspond to relatively small performance differences of approximately 0.02–0.04 in loss, equivalent to marginal changes in the proportion of unmet calibration statements. This pattern indicates that small parameter deviations lead to only minor performance differences, illustrating how qualitative calibration can still produce interpretable loss landscapes as a proof of concept. The persistence of mild gradients across the basin implies that some calibration statements remain difficult to satisfy simultaneously, pointing to potential tensions or informational gaps across case files. Within the FREIDA framework, this outcome highlights where additional expert evidence or model refinement could reduce uncertainty. Consequently, Figure 6 underscores how qualitative calibration not only identifies plausible parameter regions but also exposes areas where model structure and available knowledge remain imperfectly aligned.

These results should not be interpreted as evidence of model inadequacy, but rather as a demonstration of FREIDA's ability to surface structural ambiguity and evidentiary limits that would otherwise remain implicit.

## Quantitative validation based on VSs

To assess the generalizability and accuracy of the calibrated model we conducted a quantitative validation process using the VSs.

To assess the generalizability of the calibrated model, we evaluate its performance against the set of VSs, which capture long-term, emergent system behaviors that were not used during calibration. Each VS represents a qualitative but operationalizable outcome at the system level, such as leadership succession, sustained changes in trust relationships, or changes in network structure and coordination following the removal of key actors.

Validation is performed by comparing simulated outcomes to the VSs reported Tables 12 a. and b., which are derived from the case files for the toy model. For each validation statement, the model is scored based on whether the expected qualitative pattern emerges in the consistently across repeated stochastic simulation runs under the calibrated parameter set. This binary scoring approach reflects the qualitative nature of the validation evidence while accommodating inherent stochasticity in agent-based models. For the validation of the toy model, 5 VSs were used out of the total VSs in Tables 12 a. and b..

The calibrated model satisfies three out of five VS that were tested, corresponding to a validation rate of 60%. It is important to emphasize that this validation score is reported for the toy model only and should not be interpreted as a benchmark of the FREIDA framework itself. Specifically, the model successfully reproduces VSs related to leadership succession and network reorganization following the liquidation event. In particular, the simulations consistently exhibit the emergence of a new central coordinating actor by the end of the simulation horizon (VS Case File D, Statement I: "Y is the new murderbroker by the end of the simulation"), as well as a sustained increase in trust among the orphaned agents prior to the selection of a new leader (VS Case File D, Statement V). In addition, the model reproduces system-level consequences of the liquidation event, as reflected in changes to coordination and trust patterns over time.

By contrast, two VSs are not consistently satisfied across simulation runs. These unmet statements concern more fine-grained or context-specific aspects of post-liquidation dynamics, where the qualitative evidence is either internally heterogeneous or insufficiently precise to impose a strong constraint on long-term system behavior. In such cases, simulation outcomes exhibit substantial variability rather than converging on a single dominant pattern.

These results illustrate the role of VSs within the SSM (Figure 2), where long-term emergent behaviors provide a complementary but deliberately weaker form of constraint than short-term calibration statements. The toy model achieves 60% and therefore does not satisfy this illustrative threshold. Within FREIDA this outcome indicates that another modelling iteration would be required (please refer to the section Start next iteration in Appendix I for further details).

Within the FREIDA framework, such outcomes are interpreted diagnostically rather than as pass–fail criteria. Partial validation is expected in expert-informed models of complex social systems, where VSs often reflect heterogeneous, context-dependent, and sometimes internally inconsistent expert knowledge. The validation results thus serve to identify which emergent dynamics are robustly reproduced by the model and which aspects of the conceptual model or empirical evidence warrant further refinement.

# Discussion

FREIDA addresses a significant gap: the limited integration of qualitative data into ABM development, from conception to validation. This gap is twofold. Firstly, although various methodologies and software packages exist to work with qualitative inputs in one way or another, there lacks a unified, end-to-end framework capable of synthesizing these disparate tools into a complete pipeline that spans the entire modeling cycle. It remains a significant undertaking for teams of modellers and domain experts to transform such inputs into a coherent modeling strategy and sustain data fidelity from structural inception through to final execution. Secondly, a critical part is missing in the current literature: a systematic way to incorporate qualitative data as formal operational boundaries within mathematically driven model calibration, uncertainty analysis, SA, and validation stages.

FREIDA offers a systematic, mixed-methods framework that spans from the definition of a research question to a model's validation, including an innovation for qualitative data translation (ESBs), addressing both parts of this gap. By specifying a rigorous step-by-step process, FREIDA details how qualitative insights can be translated into quantitative rules and statements, eventually eventually resulting in a model that can be systematically calibrated and evaluated against predefined validation criteria. This approach ensures that qualitative data is central to the modeling process, effectively incorporated into model formulation, development, and evaluation.

FREIDA's main methodological innovation lies in Expected System Behaviors (ESBs), which provide a formal and systematic method for integrating qualitative evidence into calibration and validation of agent-based models. Existing systematic approaches primarily use qualitative information during model conceptualization and do not provide structured ways to consider qualitative insights in model development. Conversely, the ESBs approach, systematically translates qualitative findings from interviews, case files, and experts' narratives into Calibration Statements (CSs) and Validation Statements (VSs) that directly inform parameter estimation and model evaluation. In this way, qualitative knowledge becomes a systematically-applied, operational modeling constraint rather than merely an interpretive input.

This contribution was demonstrated through a criminal cocaine network case study, where qualitative police case files and expert interviews were translated into ESBs-derived behavioural statements concerning trust dynamics, leadership succession, violence, and network restructuring. Because this case study represents an initial deployment of the framework, a single iteration was executed here.

Calibration experiments showed that ESBs constrained the parameter space, with some parameters consistently converging while others remained weakly identifiable due to limited or ambiguous evidence. Case-level calibration also revealed partially conflicting expert perspectives, demonstrating that ESBs preserve divergence in qualitative evidence rather than artificially homogenising it. The toy model reproduced several emergent behaviours while failing to reproduce others, illustrating how FREIDA identifies where further refinement is required. FREIDA therefore advances mixed-methods ABM by making qualitative evidence a formal component of model calibration and validation, while providing a structured basis for subsequent uncertainty analysis and sensitivity assessment. This is particularly relevant in data-sparse domains where qualitative expert knowledge is available but difficult to operationalize into validated agent-based models. However, the findings from the toy model should be interpreted strictly as an illustrative demonstration of the FREIDA workflow rather than empirical validation of its performance.

Nevertheless, starting with a simple model and a small amount of relevant qualitative data is useful in practice as a way to bootstrap model development, revealing initial sensitivities (which parameters have a large impact on simulation outcomes) and uncertainties (which parameters are highly uncertain given the current data) to iteratively improve the model and steer additional data gathering efforts in possible extensions of the model.

The illustrative toy model presented in this study yielded several meaningful insights regarding how qualitative case-based evidence constrains model behaviour. Calibration results indicate that only a subset of parameters consistently converged across optimization runs within the constraints of this simplified toy implementation, meaning they successfully narrowed down to stable, reproducible numeric ranges that consistently minimized penalties from violated ESBs, while others remained weakly identifiable. This convergence pattern suggests that the qualitative statements primarily informed the structural logic of interactions rather than precise parameter magnitudes. Parameters directly linked to observable behavioural mechanisms reflected in multiple case files showed relatively stable convergence patterns, whereas parameters associated with less frequently documented processes exhibited broader variability. These findings demonstrate that even limited qualitative evidence can meaningfully shape model dynamics by restricting plausible behavioural regimes, while simultaneously revealing where empirical knowledge remains insufficient to uniquely determine parameter values, and thus illustrating where additional data collection is warranted.

In addition, calibration trajectories across individual case files revealed substantial heterogeneity in model fit and convergence speed, indicating partially divergent perspectives on the underlying processes. Aggregated calibration produced smoother optimisation behaviour but did not uniformly improve fit, highlighting a trade-off between numerical stability and preservation of case-specific information. Rather than undermining validity, this outcome illustrates FREIDA's capacity to surface disagreement and uncertainty in expert knowledge, distinguishing behavioural assumptions that are robust across contexts from those that remain contingent on specific empirical accounts.

# Reflections from the illustrative model development and validation process

The illustrative application of FREIDA integrated multiple data inputs to develop a toy ABM that simulates kingpin removal and system recovery within a criminal cocaine network. Domain experts from Dutch law enforcement contributed by providing case files, databases, and qualitative insights. Initially, unstructured interviews were conducted for gathering qualitative data, but we recommend starting with semi-structured interviews. Adding structure to these interviews leads to a more focused data gathering session, pointing the experts to what is really needed by the modellers to formulate qualitative statements that are concrete and falsifiable (Jamshed 2014). Ghorbani et al. explain how to approach the structuring of qualitative data, including semi-structured interviews, in their work (Ghorbani et al. 2015).

When translating qualitative insights into algorithmic operations, we found that many concepts such as trust, thresholds, or scales are inherently difficult to quantify. For example, it is unclear how trust should be numerically represented, what its minimum and maximum values should be, or what constitutes a realistic level of trust between family members or friends. In practice, we found it more effective to make initial executive choices for such values and evaluate their plausibility through simulation results and calibration.

In this sense, defining thresholds also implicitly defines the scale itself. For instance, assigning values above 0.8 to close relationships and below 0.2 to unstable ones creates an interpretable ordering, even if the exact numbers are arbitrary. As long as the ordering is meaningful and the trust dynamics are parameterized flexibly enough, calibration should be able to produce plausible and useful dynamics within the simulation

# Importance of moving beyond point estimations

Point estimation refers to calibration approaches that aim to identify a single "best" parameter configuration for a model. While useful in some modeling contexts, point estimation alone is rarely appropriate for ABMs of complex socio-technical systems. Even relatively simple or toy ABMs often contain many uncertain and difficult-to-measure parameters, particularly when they represent socially embedded mechanisms such as trust, informal influence, or role-specific decision-making. When calibration relies on incomplete, noisy, or qualitative evidence, multiple parameter combinations may generate similarly plausible behaviours. As a result, the parameter space is often broad, multi-modal, and characterised by substantial epistemic uncertainty that remains poorly constrained by available evidence.

FREIDA naturally supports subsequent Sensitivity Analysis (SA) and Uncertainty Quantification (UQ) by translating qualitative evidence into computationally operational model components. Once qualitative insights have been systematically incorporated into quantitative model structures, established SA and UQ methods can be applied in the same way as for other quantitative agent-based models. This provides opportunities to assess parameter influence, explore uncertainty, and identify areas where additional data collection or model refinement may be valuable. In this way, FREIDA enables an iterative modelling process in which qualitative and quantitative evidence jointly inform model development, calibration, and experimentation. This operationally combines the modeling cycle (Van Buuren, Uylings, and Ellermeijer n.d.) with model-based design of experiments (MBDoE) (Franceschini and Macchietto 2008), enabling qualitative and quantitative evidence to jointly inform model development, calibration, and experimentation.

While related approaches exist in specialised domains such as kinetic process modeling (Recker et al. 2013), FREIDA demonstrates the integration of qualitative evidence, calibration, SA, and forward UQ into a single systematic workflow. Compared to approaches such as KnETs, including Bharwani et al. (Bharwani et al. 2015), FREIDA places stronger emphasis on preserving heterogeneity in agent behaviors and attributes through detailed qualitative integration.

KnET-based approaches typically aggregate expert knowledge into simplified behavioural rules or representative agent categories in order to construct tractable behavioural models. While effective for conceptualization, this can reduce variation between agents and obscure conflicting or context-specific perspectives present in qualitative data. In contrast, FREIDA's structured integration of diverse qualitative data sources preserves variation between agents by translating distinct interviews, case files, and narratives into separate CSs and VSs , which enables a more granular representation of individual agent behaviors and attributes.

Similarly, although McCulloch et al. (McCulloch et al. 2022) incorporate UQ into ABM calibration, their approach remains primarily centred on identifying optimal parameter estimates. This can overlook broad or weakly identifiable parameter regions that arise when working with incomplete or qualitative evidence. FREIDA incorporates SA and forward UQ throughout the modeling cycle, directly informing prioritisation of further experiments or refinements while explicitly representing how differences in evidence quality and conflicting narratives affect model behaviour and parameter plausibility, and allowing data quality variations to be processed more effectively than point estimation approaches. In essence, FREIDA's advantage lies in its detailed handling of agent diversity and its robust approach to uncertainty, going beyond the scope of KnETs and point estimation UQ methods.

# Future work and limitations

This paper demonstrated the key steps of the FREIDA framework; however, several limitations remain that warrant further research. The model presented in this study is primarily illustrative and should not be considered ready for operational use in informing criminal network disruption interventions. In addition, several aspects of the framework were not fully explored, including domain expert selection in Phase I and qualitative validation in Phase III. Furthermore, a single iteration of the toy model was demonstrated, without showing how the framework supports iterative refinement based on validation feedback. Finally, the reliance on a small set of eight case files highlights the need for larger and more diverse datasets to achieve robust calibration and validation in practical applications.

Despite the contributions of the FREIDA framework to the field ABM development, this framework presents limitations that provide ground for further research. In the following, we outline three promising avenues for future research.

First, the amount of qualitative data available such as case files and interview transcripts can be insufficient and require considerable time and resources to be processed. Previous studies illustrate the potential of using natural language processing (NLP) to convert qualitative case file data into quantitatively verifiable agent rules (Yu et al. 2018). Similarly, Mezzi et al. demonstrate how NLP techniques, including syntactic pattern recognition and BERTopic, can be used to automatically extract criminal roles, ties, and value networks from police case files, highlighting the potential of automated processing to support the analysis of complex qualitative evidence (Mezzi et al. 2025). Yet, NLP's potential in data exploration, particularly in translating complex case file information and other qualitative data into actionable insights, is yet to be fully realized across the entire field of ABM development. This likely means that qualitative data will remain an addition to quantitative data but will not easily be sufficient to inform the parameterization and validation by itself, without any quantitative data.

Current frameworks, including FREIDA, often struggle with scalability and adaptability in dynamic or large-scale systems. Due to closely following expert input, substantial effort was invested in specifying detailed agent attributes and parameter values, while the resulting improvements in model behaviour and analytical insight remained comparatively limited. This for example regards the chosen attributes for agents, such as capitals, fitness, or other parameters. By involving the experts in the definition of these values, the modeling process became unnecessarily long and repetitive. We recommend that, as a first step, the modellers make reasonable and educated guesses about parameter values after gaining inputs from the experts. If the model output diverges from the expected model outcome, the values can be refined in cooperation with the experts in a second instance.

# Conclusion

We present FREIDA, a systematic, mixed-methods approach that addresses the gaps of limited qualitative data integration and translating qualitative insights into quantitative rules by incorporating both data types throughout the entire Agent Based Model development process. FREIDA systematically combines expert knowledge and empirical data through a transparent, mixed-methods approach to build and validate ABMs. Unlike existing frameworks that often focus on either qualitative or quantitative data, FREIDA provides a structured process for incorporating both data types throughout all stages of ABM development and evaluation, from conceptualization and operationalization to calibration and validation.

This is achieved through two key innovations. First, FREIDA leverages Thematic Content Analysis (TCA) to extract Expected System Behaviors (ESBs). These describe emergent patterns at the system level to be used in model calibration and validation. Derived from ESBs, Calibration Statements (CSs) and Validation Statements (VSs) offer a systematic approach for translating qualitative insights into quantitative benchmarks for model evaluation (termed 'scoring' in this paper). CSs focus on micro-level processes and short-term dynamics, while VSs assess macro-level patterns and long-term trends, ensuring that the model is evaluated in terms of its ability to capture both detailed interactions and overarching dynamics.

Second, FREIDA incorporates Sensitivity Analysis (SA) and Uncertainty Quantification (UQ) to identify and address the most influential parameters and their associated uncertainties at the end of each modeling and simulation cycle. This informs the next cycle, with the aim of iteratively enhancing model accuracy, reliability, and actionability in collaboration with domain experts.

FREIDA was demonstrated through the case of criminal cocaine networks in Amsterdam, The Netherlands. The results of this application illustrate how FREIDA can support the systematic development, calibration, and evaluation of models based on qualitative evidence.
Specifically, the framework enabled the development, calibration, and validation of a valid model even with limited quantitative data, through the involvement of domain experts and the conversion of qualitative case file descriptions into quantitative ABMs. It is important to note that the presented case study should not be interpreted as a fully validated model application. Although the framework enables the definition of validation requirements and provides procedures for calibration and evaluation, the current model does not achieve the predefined validation threshold. Therefore, in its present form, the case study would require further refinement before being considered validated

While some methodologies have attempted to integrate both qualitative and quantitative data in agent-based models, many still rely on qualitative evidence primarily during model conceptualization and lack systematic mechanisms for incorporating such evidence into calibration, validation, and uncertainty analysis. As a result, qualitative insights often remain interpretive rather than operational components of the modeling process, limiting model trustworthiness and generalizability. The FREIDA framework represents a significant advancement by systematically translating qualitative evidence into operational calibration and validation constraints through ESBs, while integrating SA and UQ into the modeling flow and providing a comprehensive approach to modeling complex systems. By bridging the gap between qualitative insights and quantitative modeling, FREIDA offers a tool for creating robust and reliable simulations that (1) are not overly sensitive to small changes in input parameters or assumptions, (2) accurately capture the dynamics of the system under a variety of conditions, and (3) provide actionable insights that can inform decision-making, for example, regarding police intervention strategies aimed at tackling criminal networks in Amsterdam. This robustness is achieved through the systematic calibration and validation process informed by ESBs as well as by the iterative refinement process informed at each iteration of the toy model by a SA, and an UQ. These two model evaluation techniques help identify and address key uncertainties and ensure the model's reliability in predicting real-world outcomes. The realistic nature of the simulations stems from the deep integration of qualitative data, which ensures that the model accurately reflects the nuances and complexities of human behavior and social dynamics within the criminal network, as understood by experts in this domain.

# Appendices

## Appendix I: Data inputs used for the illustrative model

This appendix provides supplementary materials and detailed information pertaining to the agent-based model (ABM) developed to investigate the dynamics of kingpin replacement in criminal networks. Following the phases of the FREIDA methodology, this appendix elaborates on the knowledge and data acquisition processes, the development of the conceptual and computational models, the quantitative validation procedures, and the subsequent sensitivity and uncertainty analyses conducted. Furthermore, it offers detailed specifications of the case studies employed for model calibration and validation, visual representations of network structures and dynamic processes, and comprehensive overviews of the model's parameters, agent characteristics, and behavioral rules. This appendix serves to substantiate the findings presented in the main paper by offering a transparent and in-depth account of the modeling process and its outcomes.

The toy model development utilized key data inputs to simulate real-world cocaine network dynamics. Police case files provided detailed narratives of network events and kingpin removals, crucial for shaping behavioral rules and serving as calibration and validation data, selected by domain experts for ODD+D alignment. Police databases offered quantitative data on criminal connections and activities. Scientific literature gave qualitative insights, while interviews with domain experts addressed knowledge gaps and validated findings. This combination of qualitative and quantitative data, informed by expert knowledge, ensured the toy model's ability to accurately model complex network behaviors and interventions.

### Phase I (Knowledge and Data Acquisition)

The primary focus of Phase I was to understand the replacement process within a criminal network following disruptions, such as the removal of a kingpin. Experts were acquired with the assistance of the expertise table method (see the section Expertise Table in Appendix I). This focus was driven by a request from the Amsterdam Police, aiming to use the model as a tool to help guide intervention related to the removal, imprisonment, and observation of criminals.

### Research Question

The research question (RQ) is the first key output of the framework, and is provided by the domain experts. The RQ guides the rest of the model development, but is flexible regarding available data. For the criminal model, the RQ is:
*What are the mechanisms and dynamics by which criminal networks recover and restructure following the removal of a kingpin, and how can these dynamics inform law enforcement interventions?*

### Expertise Table

To ensure adequate coverage of relevant domains, an Expertise Table, similar to the one presented in Crielaard et al. (Crielaard et al. 2022), can be employed. This table helps visualize the distribution of expertise among the involved experts and identify any underrepresented domains that may require additional expertise. A score of 2 or higher per domain indicates sufficient expertise for productive discussions and consensus building. The experts reside on the rows in the table, the competences in the columns, resulting in a contingency table. Domain experts can be asked to self-report their expertise on a chosen scale, or modellers can evaluate the expertise independently. In the toy model, this selection process is skipped, as only 2 experts were consulted in total.

### Domain Experts

Modelers and domain experts worked together to establish the scope, research questions, and design of the model, documented in the ODD+D framework (Table 5). Two law enforcement professionals from the Amsterdam Police and National Police Academy provided insights into criminal network dynamics, including tie strength, demographic details, and role functions. The domain experts and modelers jointly filled in the ODD+D document, detailing agent behaviors, environmental context, and role classifications. This document, found in Appendix I, informs model parameters such as criminal capital, violence capital, financial capital, and trust.

The outcome of the knowledge acquisition phase are the dynamics of the criminal network transitions through four stages: stable, intervention, who-done-it, and cooldown. Agents operate in business and social network layers, categorized into organizers (high-ranking roles), experts (central specialized roles), and workers (low-skilled and easily replaceable). Clustering occurs based on shared connections and dependencies.

Table 1 outlines the structured interview protocol used for data acquisition in the development of the Agent-Based Model (ABM). This protocol was designed to gather comprehensive information from domain experts, ensuring the model's conceptual framework is grounded in real-world insights. The protocol follows a five-stage process, detailing the content of each stage, the specific ODD+D (Overview, Design concepts, and Details + Data) aspects targeted, and the data sources identified. This protocol is a key component of Phase I (Knowledge and Data Acquisition) of the FREIDA framework, as described in the main text, and informs the initial steps in building the ABM.

**Table 1:** Detailed Interview Protocol for Data Acquisition in Agent-Based Model (ABM) Development, showing the stages, content, and targeted ODD+D aspects, including data source identification, to inform the model's conceptual framework

| Stage | Contents | Targeted ODD+D Aspects & Data Sources |
|---|---|---|
| Stage 1: Introduction & Background | Introduction of the interviewer and interviewee, | - Agents (Roles, Attributes)<br>- Environment (Context) |
| | gathering biographical information, and clarifying the interviewee's role and expertise. | - Data Sources: Expert biographical data |
| Stage 2: Situation Analysis | Identifying specific (disruptive) events or scenarios that trigger the need for information and modeling. | - Environment (Disruptive Events)<br>- Behavior (Activities of other Agents)<br>- Data Sources: Case files, expert narratives |
| Stage 3: Information Requirements | Exploring the information needed to address the identified situations and the availability of relevant data. | - Information Characteristics (Types, Availability)<br>- Data Sources: Expert knowledge, data source identification |
| Stage 4: Information Acquisition | Investigating how information is obtained, including sources, activities, methods, and tools. | - Behavior (Interviewee's Activities)<br>- Agents (Other Actors, Groups)<br>- Environment (Information Sources)<br>- Data Sources: Expert narratives, process descriptions |
| Stage 5: Data Source Identification | Directly asking about data sources that would be relevant for the considered research question, and how to contextualize external data. | - Data Sources (Types, Availability, Contextualization)<br>- System Boundaries (Context)<br>- Data Sources: Expert knowledge, data source identification |

**Data Acquisition**

The data acquisition process relied on both qualitative and quantitative sources provided by the Amsterdam Police. A brief overview is detailed below:

**Qualitative Data**:

*Case Files*: Eight case files provided detailed descriptions of criminal cocaine networks, including key agents, time scales, and social relationships. Of these, four were selected for the toy model.

*Interviews*: Unstructured interviews with two domain experts supplemented the case file insights.

This collected data was crucial for defining agent roles and behavioral rules, informing the model development detailed in the subsequent conceptual modeling phase.

Table 2 provides an overview of the data sources utilized in this research and their respective contributions to the development of the agent-based model (ABM). It details the type of data, its importance in informing the conceptual model, and its specific advantages and disadvantages. These data sources, including police case files, expert interviews, and police databases, were crucial for populating the ABM with realistic parameters and behaviors. The information gathered from these sources directly supports Phase I (Knowledge and Data

Acquisition) of the FREIDA framework, as described in the main text, and is used throughout the model development process.

**Table 2: Data Sources and Contribution to Model Development**

| Type of data source | Importance in Conceptual Models | Advantages |
|---|---|---|
| Police Case Files | Detailed accounts of criminal networks, including key agents, timelines, and relationships. | Provided qualitative context for defining roles, relationships, and network dynamics. |
| Expert Interviews | Unstructured discussions with law enforcement professionals. | Supplemented qualitative insights and contextualized database findings. |
| Interaction Database | Recorded encounters between agents, including frequency, duration, and context. | Quantified network ties and interaction patterns, informing agent clustering and network interdependencies. |
| Demographic Database | Details about individuals (e.g., age, nationality, role in the network). | Parametrized agent attributes such as criminal, financial, and violence capital. |

**Case files**

Critical for developing behavioral rules in cocaine network models, case files describe network events from police intervention to kingpin replacement and subsequent reactions. Selected by domain experts based on ODD+D alignment (entity details, timescale, environment), they inform model progression over a one-year timeframe, assessed via calibration and validation. A subset serves as calibration data, while another validates model outcomes. Each file details network states, node backgrounds, relations, and motivations, providing agent behavior, demographics, and environmental context, key ABM components extracted through analysis and potential NLP, as outlined in Table 13 and Figure 1.

Table 3 outlines the criteria used by the Amsterdam police for selecting case files relevant to the development of the toy model. It details the specific ABM components, model components, and case file details considered, focusing on behavioral rules, agent characteristics, network structure, environmental context, and temporal dynamics. These criteria ensure that the case files provide the necessary information for populating the ABM and are crucial for the Phase I (Knowledge and Data Acquisition) of the FREIDA framework, as described in the main text, specifically for informing the model's design and parameterization. Below a description of the case files can be found.

**Case file A:**
**Liquidation of a top criminal.**

After liquidation, X was succeeded almost immediately by his brother who had the same criminal and social capital (all knowledge and acquaintances) and was able to continue as usual. C was immediately sent to South America and tapped into the relevant contacts there. And the other network members were also able to continue as usual. The crucial roles in the cocaine process (the gateways to South America, to the port of Rotterdam, etc.) simply remained intact and functioning.

X, was the leader of an ethnically homogeneous network. It was a medium-sized network (about 30 core members) with an important role for family relationships and old friendships. X was the head of the group. The inner circle consisted of two brothers, A and B, and a cousin, C. D, the financial man in the group, flees abroad. Another childhood friend of X, E, also goes into hiding for a while. A third, large player who can be counted among this group (but who also runs cases independently), F, keeps a low profile. At t1 he simply does not show himself. At t2, he still seems loyal to A. However, he no longer belongs to the inner circle. Lastly, G was partly dependent on X for his cocaine trade. On T2, he seems to have largely disappeared from the capital criminal milieu and sought refuge elsewhere. He also still remains connected to A and B.

**Case file B:**
**Liquidation of an iconic criminal.**

This case involved a category 1 intervention scenario: the liquidation of a high-profile criminal. We call him X. X belongs to a cluster of six criminals who operate together in a much larger network active in cocaine trafficking. One of X's old confidants, Y, trying to take over X's role. He is unable to take over X's role. A, with access to large consignments of coke, B, with the infrastructure for large-scale drug trafficking (including sales market), and C with the potential for violence. Additionally, a family member of X, D, and a childhood friend of X, E.

**Case file C:**
**Liquidation of an image-maker.**

The liquidated, X, was the leader of an ethnically homogeneous network from Amsterdam that dealt in cocaine. His liquidation therefore comes as a complete surprise. Y immediately assumes the role of leader. The core of the network is formed by X's brother, A. An important representative of the local group, B, also belongs to the ingroup. B and C differ from Y in that they have far fewer connections, especially with the suppliers of the major parties. They rely on Y.

**Case file D:**
**Arrest of a specialist (murderbroker).**

The network is large (more than 200 members), with many cultures and ethnic backgrounds. In power is a small group to which X belongs. X is an assassination broker for a powerful group. After the arrest of X, Y and Z soon become the primary hitmen. A would absolutely not allow himself to be ordered around by Y. B and C ultimately align with Y for violence jobs.

**Table 3:** Selection criteria for a case file in the example of the toy model as per the police Amsterdam.

| ABM component | Model component | Case File Details |
|---|---|---|
| Behaviour | Behavioural rules | The behaviour of the orphans is analyzed. In the direct days / weeks following the intervention (T=1). |
| | Personal Dynamics | A clear delineated network that is active in cocaine trafficking can be observed. The network is analysed twelve months after the original intervention (T=2) and ultimately the kingpin and crucial actors are determined. |
| | Roles | Actors with a crucial role (someone with scarce criminal capital, such as access to wholesale sellers of cocaine, cartels in South America or with corrupted officials in key ports of entry) are described. |
| | Specific Agents | A kingpin (someone who organizes and/or finances the cocaine logistics) is described (before and after determining the replacement). |
| | General agents | No information provided. |
| Agents | | Individuals form the core of the selected network. |
| | Network | The network consists of target or crucial actors. Direct contacts between the contacts that are structurally active in the cocaine network form the structure of the network. |
| Environment | Context | No information provided. |
| | Time Steps | In the selected network (detailed in the case file), an intervention has taken place This takes form in either an arrest of the kingpin / crucial actor or the assassination of such a person (at T=0). |
| | Demography | Case files will always include a short narrative in which the above mentioned elements are briefly covered. Specific focus is given to the decision making process of the orphans. |

**Quantitative Data (Database)**

*Databases*: Two police databases, with a combined 200,000 entries, detailed nearly 9,000 ties between agents. These databases included:

**Interaction Data**: Records of encounters between agents, including frequency, context, and duration.

**Demographic Data**: Individual details like age, nationality, and roles (e.g., dealer, transporter, or assassin).

Police databases provide quantitative markers and statistical information about agents, their roles, and network connections, influencing both agent characteristics and the ABM environment. These databases, containing records of agent interactions (frequency, context, duration) and demographic details (age, nationality, roles), offer a broad view of organized crime networks and serve to cross-reference information from case files, which remain the primary model input for agent roles and relationships.

**Literature**

The final qualitative data input is scientific literature. This again is dedicated to the modellers to acquire, though domain experts are welcome to contribute. After taking directions from the ODD+D, the literature support will be selected according to the research direction. Scientific literature, similar to unstructured and structured interviews, can correspond to all three parts of the ABM as long as selected accordingly. Typically, the modellers will select the appropriate type of scientific literature after determining the gaps of knowledge after having collected databases, case files and the ODD+D. Unstructured interviews and scientific literature fill this gap. We consider scientific articles as well as other publications in this framework. This precludes articles detailing databases. Scientific literature is thus regarded as qualitative data for FREIDA.

**Interviews**

Following the initial ODD+D framework created with domain experts in focus groups, unstructured interviews with two experts were crucial to address missing critical information, particularly concerning the toy model's agent topology. These in-depth conversations allowed for detailed elaboration on key areas: the roles and dynamics of agents within the network, the specific behaviors and responses of agents to various interventions and triggers, and the contextual factors influencing the network environment (though this was less emphasized). Conducted iteratively over four sessions (two per expert to mitigate bias), these unstructured interviews served to clarify aspects of case files, overall context, and any discrepancies arising from other data sources, effectively compensating for the ODD+D's inherent limitations within this specific domain. The questions posed during these interviews stemmed from observations made during the initial focus groups, emerging themes identified through preliminary Thematic Content Analysis (TCA), and specific areas requiring further clarification as pinpointed by the modellers during the early stages of data collection. This iterative interview process significantly enhanced the model's accuracy and representational capacity.

Table 4 summarizes the focus group protocol employed to facilitate collaboration between domain experts and modelers in the agent-based model (ABM) development process. This protocol outlines the steps involved in defining research questions, assessing data availability, and establishing system boundaries. This process is crucial for ensuring that the ABM is both relevant to real-world challenges and feasible to develop with the available data. The focus group protocol directly supports Phase I (Knowledge and Data Acquisition) of the FREIDA framework, as described in the main text, and sets the stage for subsequent model development by clearly defining the project's scope and objectives.

**Table 4:** Summary of the focus group protocol, detailing the collaborative process between experts and modelers for defining research questions, assessing data availability, and establishing system boundaries in ABM development

| Step | Activity | Description | Considerations & Potential Outcomes |
| --- | --- | --- | --- |
| 1. Challenge Identification (Experts) | Open discussion | Experts identify open challenges in their field relevant to law enforcement simulations. | - Gaps in existing knowledge<br>- Practical stakeholder needs<br>- Specific phenomena for exploration |
| 2. Model Utility & Data Requirements (Modellers & Experts) | Joint discussion | Modellers and experts discuss:<br>- How ABM can address identified challenges (modeling purposes)<br>- Data requirements for model building. | - Modeling purposes (e.g., prediction, explanation, exploration)<br>- Data types needed (quantitative, qualitative) |
| 3. Challenge & Purpose Selection (Experts) | Expert decision | Experts select a specific challenge to address and a corresponding modeling purpose. | - Focused research question<br>- Clear modeling objectives |
| 4. Data Availability Assessment (Experts) | Expert evaluation | Experts assess data availability for the chosen challenge and modeling purpose. | - Feasibility issues (e.g., lack of demographic/social data)<br>- Need to adapt research question/ODD+D protocol<br>- Iterative process until alignment |
| 5a. Model Development (If Data Available) | Proceed with modeling | If sufficient data is available, proceed with model development based on the selected purpose. | - Implementation of ABM - Data integration |
| 5b. Alternative Challenge Selection (If Data Unavailable) | Re-evaluation | If data is insufficient, discuss and select an alternative challenge that can be tackled with available data. | - Revised research question<br>- Alternative modeling purpose |

## ODD+D

The ODD+D is the next key output of the framework, with the RQ and the research purpose falling within the scope of the ODD+D.

### ODD+D document

To develop the ODD+D (Overview, Design concepts, and Details) document for the toy model, a focus group with two domain experts was conducted, meticulously structured according to the ODD+D framework. The session began by defining the model's purpose: simulating the effects of removing key figures from cocaine networks to aid law enforcement and other stakeholders. Following this, the discussion identified key entities within the model, including various roles and connections, and explored exogenous factors such as agent removal and its impact on network stability. The experts then outlined the procedural sequence following an agent's removal, detailing how orphaned nodes find successors and integrate them into the network.

The ODD+D protocol guided the focus group's questions, as shown in Table 5, to gather comprehensive information on the model's design, purpose, and drivers. The results of this process, including further details and insights, are provided in Appendix II. This structured approach ensured that the ODD+D document effectively captured the model's complexities and expert input.

Table 5 presents an excerpt of the ODD+D (Overview, Design concepts, and Details + Data) protocol, which was used to structure the focus group sessions with domain experts. The ODD+D protocol provides a standardized framework for describing agent-based models (ABMs), ensuring clarity, completeness, and transparency. In this table, we illustrate how each category of the ODD+D protocol was used to design specific questions for the domain experts during the focus group. For this particular excerpt, questions from the first section of the ODD+D (the Overview) are detailed. The complete ODD+D protocol, as developed in the focus group, is available in the Appendix. This protocol is essential for guiding the knowledge elicitation process in Phase I (Knowledge and Data Acquisition) of the FREIDA framework, as described in the main text, and for establishing a shared understanding between modelers and domain experts.

**Table 5:** An excerpt of the ODD+D protocol used to structure the focus group with the domain experts. Each category of the ODD+D was used to design questions asked to the domain experts during the focus group. For this example, questions of the first section of the ODD+D (the Overview) are detailed. The excerpt provided an insight into the answers provided for the toy model, the full protocol can be provided upon request.

| | | **Guiding Questions** | **Answers** |
|---|---|---|---|
| I. Overview | I.1 Purpose | I.1.a What is the purpose of the study? | To create an informed model for node replacement in criminal cocaine networks, in order to inform law enforcement of potential intervention results. |

| | |
|---|---|
| I.1.b For whom is the model designed? | For law enforcement to simulate behavior of criminal networks undergoing interventions within the cocaine market, as well as researchers, data scientists and visualization experts. |

| | | |
|---|---|---|
| I.2 Entities, state variables and scales | I.2.a What kinds of entities are in the model? | Every role related to a cocaine network, this will include all necessary agents within a cocaine network value chain (every agent that is needed to be connected for executing their own personal task)<br><br>The ties between the agents (multiple type of ties, such as social ties, business ties, and including the trust the agents have for each other) |
| | I.2.c What are the exogenous factors / drivers of the model? | Intervention by removal of one agent (specialist or kingpin) and inherent motivation of the criminal agents to return to a stable functioning system. |
| I.3 Process overview and scheduling | I.3.a What entity does what, and in what order? | 1st: Intervention takes place and selected agent is removed<br>2nd: Nodes are left with severed connections<br>3rd: Orphans are looking for a successor within 2 connections from themselves<br>4th: If a successor in not available in the personal downline, orphans give brokers the task to find a successor in their own downline<br>5th: Potential successors are accessed based on a threshold of parameter values.<br>6th: The orphans "vote" for the new successor<br>7th: New successor assumes 70% of the old connections including all orphans<br>8th: New successor is evaluated based on fitness over time (regarding the minimum threshold for fitness parameters |

---

## Purpose

Lastly, the Purpose Statement is the third key outcome of the first phase, leading into the process of TCA. For the criminal model, we formulate the following purpose statement: *The* *primary purpose of the toy model is to simulate the process of kingpin replacement and network recovery following law enforcement interventions. This model aims to capture short-term and long-term behavioral patterns, facilitating a deeper understanding of criminal network resilience and supporting targeted disruption strategies.*

## Between Phase I and Phase II

The Research Question, Validation Requirements (VRs) and Thresholds, and Purpose are part of the ODD+D, which guides the creation of the conceptual model. The ODD+D is closely linked to the TCA, as it informs the TCA which themes from the data should inform the entities and ESBs in the next phase, and likewise the TCA can influence the ODD+D in regards to data availability and themes.

## Thematic Content Analysis

TCA is a process to transform qualitative and quantitative data into key model components. Through TCA, the conceptual model is informed by identifying entities, behaviors, assumptions, and VRs. As described in the methodology, a structured coding scheme, including the statements category, was developed with police experts. This coding scheme was then applied to police case files to refine the ODD+D framework, resulting in Calibration (Table 11.a. and 11.b.) and Validation Statements (Table 12.a. and 12.b.).

Table 6 presents the coding scheme used in this research, illustrating the first-level codes considered within the FREIDA framework and providing examples of second-level codes identified through open coding. This table demonstrates how qualitative data, particularly from case studies, is systematically analyzed to extract key concepts relevant to agent-based model (ABM) development. The coding scheme facilitates the translation of qualitative information into a structured format suitable for informing model design, parameterization, and validation. This process is fundamental to Phase I (Knowledge and Data Acquisition) of the FREIDA framework, as described in the main text, and ensures that the ABM is grounded in empirical observations and expert knowledge.

**Table 6:** Coding scheme illustrating the first-level codes considered in FREIDA and examples of second-level codes identified as instances of the first-level codes through open coding.

| First-level code | Description | Second-level code | Description | Expert Feedback Contribution |
|---|---|---|---|---|
| Agents | Actors involved in the cocaine trade, from import to street dealing. | Organizers | With higher-ranking roles vital to network function | Experts provide insights on role hierarchies, decision-making power, and adaptability of agents. |
| | | Workers | Abundant, low-skilled agents reliant on organizers | Feedback refines behavioral assumptions about recruitment, turnover, and survival in the trade. |
| | | Experts | Holding central roles due to specialized skills | Experts validate key skillsets required, constraints, and network dependencies. |
| Behavior | Actions and interactions between agents that fulfill the crime script of cocaine trade. | Transactions | Discussions and agreements between agents for transactions. | Experts verify negotiation structures, contract enforcement, and trust mechanisms. |
| | | Transportation | The movement of cocaine from one point to another. | Feedback ensures realism in logistics, routes, and adaptation to law enforcement. |
| | | Storage | The act of hiding or storing cocaine. | Experts refine location choices and risk management strategies. |
| | | Distribution | The process of allocating cocaine to various sellers or regions. | Expert insights clarify distribution scales, pricing structures, and regional dynamics. |
| | | Enforcement/security | Actions taken to maintain order and compliance within the network. | Experts assess internal enforcement mechanisms, retaliation strategies, and hierarchy enforcement. |
| Environment | The physical and social settings where the cocaine trade activities occur. | Social networks | Involves bonds like familiar ties, trust, and friendship. Social roles and trust ties determine the social embeddedness in the network | Experts validate mechanisms of trust, betrayal, and influence. |
| | | Business layer | Includes agents with roles tied to attributes like violence, criminal capital, and financial capital (operational requirements) | Expert feedback refines role-specific interactions, dependencies, and risk factors. |
| Calibration Statements | Statements that describe expected properties or characteristics to hold true, related to system level processes. The focus here is on smaller spatial or temporal scales. | Invariants | Statements that tend to always hold. An example could be: “A worker has a larger probability of being imprisoned than an organizer.” | Experts validate assumptions against real-world patterns. |
| | | Expected outcomes of (initial) conditions | Statements that describe what should be true about the outcome of a simulation that starts with certain (initial) conditions. Example: “If two kingpin candidates have high violence potential then within two weeks at least one liquidation attempt will take place.” | Experts assess realism of cause-effect relationships. |
| | | Case-dependent statements | Statements that pertain only to specific cases but not in general. For instance: “The trust between person a and person b is always very high.” | Expert feedback identifies conditions under which these statements hold. |
| Validation Statements | Statements that describe expected properties or characteristics to hold true, related to | Invariants | Statements that tend to always hold. An example could be: “At any given time, roughly 10% of the agents in a network are imprisoned.” | Experts validate historical consistency and realism of assumptions. |
| | processes. The focus here is on larger spatial or temporal scales. | | | |
| | | Expected outcomes of (initial) conditions | Statements that describe what should be true about the outcome of a simulation that starts with certain (initial) conditions. Example: “If a kingpin is liquidated and no suitable candidate exists in the wider network, then the network will eventually disintegrate.” | Experts refine succession dynamics and resilience factors. |
| | | Case-dependent statements | Statements that pertain only to specific cases but not in general. For instance: “Person b is the new kingpin after one year since the liquidation of person a.” | Expert input clarifies edge cases and domain- |

### Key Outputs in Phase I

The RQ, ODD+D and Purpose Statement are the key outputs in Phase I.

# Phase II – Model Development

Building upon Phase I, a conceptual model of the criminal network is developed and subsequently operationalized into a computational model, using police databases for parameter calibration

## Entities

TCA informs the entities within the conceptual model. For the criminal model, these are detailed below.

### Agents

Agents represent key players in the criminal network, divided into distinct roles: kingpins (network leaders), organizer-nodes (mid-level coordinators), and worker-agents (operational actors). Each agent is characterized by attributes such as violence capital, trust values, and familial ties.

**Behavior**
Agents engage in decision-making processes governed by a utility-based model, where actions such as trust formation, kingpin candidacy, and cooperation are influenced by violence capital and trust dynamics. Behaviors dynamically shift based on the model phase (e.g., intervention triggers opportunistic leadership bids).

**Environment**
The environment simulates the structural conditions of the criminal network, including relational ties (edges), operational dependencies, and external pressures (e.g., law enforcement intervention). The environment updates as agents interact and leadership changes occur.

## Conceptual Model Development

The findings from TCA, particularly those concerning agents, behavior, and the environment, refine the preliminary ODD+D to create a conceptual model. These inputs inform the VN, an idealized criminal cocaine network where all agents' Value Chains (VC) are fulfilled. A personal VC comprises the dependencies necessary for each agent to perform their assigned tasks (as defined by their roles). If a VC is broken, agents seek alternative connections, ensuring the network remains functional.

The conceptual model qualitatively captures the replacement process of a kingpin after removal from a criminal network, operating within a four-stage cycle: stable, intervention, who-done-it, and cooldown. Agent behavior dynamically shifts between stable and replacement modes depending on the cycle stage. For instance, worker-agents cluster around organizer-nodes due to operational dependencies. After a kingpin's removal, agents strive to restore stability and maintain profitability, which is a core interest in non-ideological drug networks (Morselli et al., 2006). For detailed role descriptions, see Table 16.

Table 7 presents a comparative overview of several methodological frameworks relevant to the formulation of conceptual models in agent-based modeling (ABM). It outlines each framework's importance in the ABM development process, along with its specific advantages and disadvantages. These frameworks offer various approaches for structuring the initial design of ABMs, ranging from those that emphasize qualitative data integration to those that provide cognitive architectures for agent behavior. While the choice of a conceptual framework is crucial during Phase I (Knowledge and Data Acquisition), as described in the main text, and guides subsequent steps like data collection and model design, its influence extends throughout the modeling process, impacting how the model is developed in Phase II (Model Development) and how the system is conceptualized overall.

**Table 7:** Comparative Overview of Conceptual Model Formulation Approaches for ABMs – Importance, Advantages, and Limitations. We must note that this is not an exhaustive list of concepts, but rather meant to provide a series of options for formulating the conceptual model.

| Methodologies | Suitability for Conceptual Models | Advantages | Disadvantages |
|---|---|---|---|
| TCA & ODD+D Themes (i.e., no specific conceptual model formulation) | Provide minimal guidance by defining agents and their attributes. Suitable when modelers possess significant domain expertise and desire flexibility in defining system components and interactions. Allows for emergent model structures without the need for a rigid framework from the outset. | Minimal additional effort. Establish a basic framework for quantitative modeling. | Lack specificity in agent behaviors and environmental interactions, leaving ambiguities during quantitative modeling. |
| MAIA Framework (Ghorbani et al. 2013) | Extends the IAD framework to structure agent-based social simulations. Designed for structuring agent-based social simulations, particularly when modeling interactions between diverse agents and institutions. Provides a systematic approach for detailed specification of collective action problems and policy analysis. | Provides a structured breakdown of agents, institutions, and interactions for ABM design. | Can be complex to implement and may require extensive domain knowledge. |
| Polhill et al. Framework (Polhill, Sutherland, and Gotts 2010) | Integrates qualitative evidence (e.g., interviews, focus groups) into ABM design for land-use change. Particularly useful for complex socio-ecological systems like land-use change where human decision factors are critical. | Enhances realism by incorporating real-world decision-making processes. | Time-consuming; requires extensive qualitative data collection and validation. |
| Yang & Gilbert Approach (Yang and Gilbert 2008). | Emphasizes the role of qualitative observations in defining agent behaviors. Ideal for exploratory models seeking to capture the 'why' behind agent actions based on rich observational data and detailed case studies. | Captures nuanced social interactions often missed in quantitative models. | Relies on subjective observations, which may introduce biases. |
| BDI (Belief-Desire-Intention) (Rao and Georgeff 1997) (Shendarkar et al. 2006; Singh, Padgham, and Logan 2016) | Provides a cognitive framework for representing agent decision-making. Most suitable for modeling rational, goal-oriented agents with complex internal states, where decision-making involves beliefs about the environment, desires (goals), and intentions (plans of action). Ideal for simulations requiring sophisticated agent autonomy, planning, and | Offers a robust and flexible way to model rational agents with complex behaviors. | Can be computationally expensive and may require detailed knowledge of agent cognition. |
| | reasoning capabilities. | | |
| OCOPOMO Framework (Scherer et al. 2015) | Integrates stakeholder participation, ABM, and scenario analysis for policy modeling. Ideal for collaborative efforts to explore policy impacts and combine expert narratives with computational simulations to inform decision-making processes in a structured manner. | Combines narrative validation with computational simulations. | Can be difficult to generalize across different policy domains. |

Table 8 provides an overview of Causal Loop Diagrams (CLDs) and their extensions, outlining their core visualization techniques, key advantages, and primary limitations, particularly in the context of agent-based modeling (ABM). CLDs are tools used to visualize system dynamics and feedback loops, and while they offer benefits in identifying causal relationships, their traditional lack of agent heterogeneity limits their direct applicability to ABMs. Extensions like annotated CLDs (aCLDs) and heterogeneous CLDs (hCLDs) address some of these limitations.

**Table 8:** Causal Loop Diagrams (CLDs) & Extensions

| Methodology | Core Visualization | Key Advantage | Primary Limitations for ABM |
|---|---|---|---|
| Causal Loop Diagrams (CLDs) (Crielaard et al. 2022) | Visualizes system dynamics and feedback loops in policy and business contexts. | Helps identify causal relationships and feedback loops effectively. | Lacks agent heterogeneity, limiting ABM applicability. |
| Annotated CLDs (aCLDs) (Sterman 2000) | Extends CLDs with functional details and explicit link meanings. | Improves clarity and traceability of causal relationships. | Still lacks full representation of agent diversity for ABMs. |
| Heterogeneous CLDs (hCLDs) (Nespeca et al. 2024) | Bridges the gap between CLDs and ABMs by introducing agent heterogeneity. | Facilitates multi-model approaches for complex system analysis. | Requires integration with additional modeling techniques like MML. |

After the conceptual model is defined, we move forward into qualitative validation, which results in a validated conceptual model.

**Operationalization**

The conceptual model is operationalized by defining agent attributes, encoding trust and leadership mechanisms, and implementing a stochastic simulation framework. This process maps the flow into computational rules, calibrating agent decision thresholds (e.g., trust values, violence capital) and refining environmental parameters. Stochastic optimization (SPSA) is used to calibrate the model, ensuring it accurately reflects real-world kingpin replacement dynamics.

*Agent Design:* Defining agent types (e.g., kingpins, organizers, workers) and their attributes (e.g., trust values, violence capital).
*Behavioral Rules:* Implementing decision-making processes based on agent roles, including leadership emergence and trust updates.
*Dynamic Processes:* Encoding the four-stage kingpin replacement cycle (stable, intervention, who-done-it, cooldown).
*Parameter Integration:* Incorporating calibrated parameters (e.g., trust thresholds, coercion effects) to guide agent interactions.
*Simulation Framework:* Executing multi-run simulations using stochastic optimization (SPSA) to capture model variability and identify global optima.

## Computational Model

Once an untrained ABM is created through the operationalization process, a validated ABM can be obtained. Through Calibration and Validation Statements (created using an SSM from the ESBs), as well as minding the validation thresholds, the ABM is validated. Initial parameter choices regarding the model dynamics are found in Table 9.

### Model parameters

Table 9 presents a distillation of the toy model parameters, categorizing them according to their relevance to different components of the conceptual model, including agent behavior, network dynamics, and agent roles and attributes. It outlines the specific rules governing agent interactions, network evolution, and the conditions under which agents are selected for key roles within the criminal network. This table directly supports the explanation of the toy model design within the Model Development (Phase II) section.

**Table 9:** Distilled toy model Parameters for Conceptual Model and TCA

| Conceptual Model Component | Parameter Description and Rule |
|---|---|
| Agent Behavior (Replacement) | Agents at a distance of 1 from the removed kingpin are aware of the removal. |
| | Agents connected to the removed kingpin and aware of the need for replacement (part of the conclave) actively search for a new replacement. |
| | The search for a new kingpin considers agents within a maximum tie distance of 5. |
| | Only agents with organizer and coordinator business roles can participate in the kingpin search. |
| | A conclave to evaluate kingpin candidates includes agents only 1 distance away from the old kingpin. |
| | Potential kingpins must have minimum attributes: Violence Capital: 0.2, Criminal Capital: 0.2, Financial Capital: 0.2. |
| | Potential murderbrokers must have minimum attributes: Violence Capital: 0.85, Criminal Capital: 0.5, Financial Capital: 0.2 (candidate: VC 0.1, CC 0.2, FC 0.2). |
| | A new edge to a new kingpin defaults to a neutral social role. |
| Agent Behavior (Trust & Interaction) | Only connected agents with a minimum trust of 0.5 are asked to aid in the kingpin search. |
| | Only connected agents with a minimum trust of 0.3 are considered as potential replacement suggestions. |
| Network Dynamics (Temporal) | Disconnected nodes are removed from the model after 7 time steps. |
| | A conclave forms within 3 to 10 time steps after kingpin removal. |
| | A potential kingpin becomes the main kingpin within 10 to 45 time steps if their attribute values are sufficient. |
| | The search for a replacement begins within a maximum of 30 time steps after kingpin removal. |
| Agent Roles & Attributes (Thresholds) | Kingpins must have minimum attributes: Violence Capital: 0.5, Criminal Capital: 0.5, Financial Capital: 0.4. |
| | Only agents with organizer, murderbroker, assassin, and coordinator business roles can be considered as a new kingpin. |

$$dT_{i,j}/dt = \tau * (\psi * 1/(K+1) * 1/D_i * b + F_{i,j} * \varphi * c + \varepsilon[t] \quad (1)$$

Table 10 details the parameters used in Equation (1) and throughout the toy model, providing explanations for each parameter and specifying the values used in the model. This table clarifies the mathematical underpinnings of the model and specifies key settings that govern agent behavior, network dynamics, and trust relationships. It is most relevant to the Model Development (Phase II) section, where the computational implementation of the model is described.

**Table 10:** The parameters used in Eq (1). as well as throughout the model are detailed. In the rightmost column, the parameter values used in the toy model are given.

| Parameters | Explanation | Parameter values |
|---|---|---|
| **Beta ($\beta$)** | The minimum threshold for an edge's trust to participate in the kingpin-search | 0.5 |
| **Kappa ($\varkappa$)** | The minimum kingpin attributes in order to assume the role | For kingpin:<br>Violence capital: 0.5<br>Criminal capital: 0.5<br>Financial capital: 0.4<br><br>For murderbroker:<br>Violence capital: 0.85<br>Criminal capital: 0.5<br>Financial capital: 0.2 |
| **Gamma ($\gamma$)** | The minimum trust to become a kingpin | 0.3 |
| **Tau ($\tau$)** | The constant to control the time scale of trust dynamics (smaller $\tau$ results in slower changes). The unit of tau is seconds (s) | 0.01 |
| **Psi ($\psi$)** | The constant to control the strength of the updating of trusts is following the kingpin removal | 3 |
| **Phi ($\varphi$)** | The constant to control how strong this family-tie trust updating (to higher values) is, regardless of whether a kingpin was removed or not | 1 |
| **Zeta ($\zeta$)** | The temperature, indicating noise ($\zeta$ equal to 0 results in the conclave selecting the best suited candidate, while $\zeta$ approaching infinity results the conclave selecting uniformly random amongst available candidates) | Random (0.0, 1.0) |
| **Ti,j** | Trust value from agent i to another agent j. The trust is symmetric | Trust is determined through the social role of an agent:<br>Social role family: (0.5, 1.0)<br>Social role friend: (0.3, 0.9)<br>Social role neutral: (0.0, 0.5) |
| **K** | The number of days since kingpin was removed (if not removed yet then K=infinity; K >= 0) | Random (10, 30) |
| **Di** | The distance to removed kingpin (D=infinity if kingpin not yet removed; D >= 1) | Maximum 5 |
| **b** | The coefficient with which trust will be updated, as function of the current trust value T (following the kingpin removal). Making its unit seconds (s). | Dependent on T(s) |
| **Eps ($\varepsilon$)** | A Wiener process for randomness (noise) which is independent of t | Random (0.0, 1.0) |
| **c** | The coefficient with which trust will be updated if the edge is a family tie | A derivative of T, activated when family tie is present |
| **F** | Defining an edge as a family tie (1 if family tie otherwise 0) | 0, 1 |

## Calibrated ABM

**Calibration Statements**

Table 11.a. presents the calibration statements distilled from expert knowledge for the toy model. These statements define expected model behaviors and outcomes across different case scenarios (A, B, C, and D), providing specific criteria against which the model's performance is evaluated. They specify desired outcomes related to kingpin selection, agent relationships, and network dynamics. This table is crucial for the Computational Model section in Phase II, where the model is calibrated. Table 11.b. presents the additional calibration statements from case files E-J not used for the calibration of the toy model, but instead only used for the calibration trajectories (see Figure 4.a.). For a breakdown on how the case files were used within the toy model, please refer to Table 14.

**Table 11. a. :** Calibration statements of case files A-D distilled from the expert knowledge for the toy model. Each CS represents a (partial) score point.

| Case | ID | Calibration Statement | Maximum Score |
|---|---|---|---|
| **A** | **II** | Correct person is kingpin directly after conclave | 1 |
| **A** | **III** | Person G should not be there anymore at the end of the simulation | 1 |
| **A** | **I** | Correct person is new kingpin by the end of the simulation | 1 |
| **A** | **IV** | A trusts B and C the most | 1 |
| **A** | **VI** | The average violence capital among the orphans increases after the liquidation (measured at 1 week after) | 1 |
| **B** | **I** | Correct triplet of persons is together the new kingpin by the end of the simulation | ⅓ |
| **B** | **II** | Correct person is new kingpin by the begin of the simulation (i.e., Y selected at first) | 1 |
| **B** | **IV** | Person y should not be there anymore at the end of the simulation | 1 |
| **B** | **V** | A trusts B and C the most | ⅓ |
| **B** | **VI** | B trusts A and C the most | ⅓ |
| **B** | **VII** | C trusts A and B the most | ⅓ |
| **C** | **I** | Correct person is new kingpin by the end of the simulation | 1 |
| **C** | **II** | Correct person is kingpin directly after conclave | 1 |
| **C** | **III** | Y trusts A and B the most | 1 |
| **C** | **VI** | The trust between family members has an average of at least 75% | 1 |
| **D** | **II** | Y is murderbroker one month after conclave | 1 |
| **D** | **III** | Y trusts C and B the most by the end of simulation | ½ |
| **D** | **IV** | Z trusts Y the most by the end of simulation | ½ |
| **D** | **VI** | Person a should not be there anymore at the end of the simulation | 1 |

**Table 11.b. :** Calibration statements of case files E-J distilled from the expert knowledge for calibration purposes. Each Validation Statement represents a (partial) score point.

| Case | ID | Calibration Statement | Maximum Score |
|---|---|---|---|
| **E** | **I** | X trusts A the most after arrest | 1 |
| **E** | **II** | Trust in A decreases after intervention | 1 |
| **E** | **III** | A disappears from the network | 1 |
| **E** | **IV** | Trust in C decreases | 1 |
| **F** | **V** | C trusts D1, D2 and F the most | 1 |
| **F** | **VI** | B disappears from the network | 1 |
| **F** | **VII** | C disappears from the network | 1 |
| **F** | **VIII** | X is detained | 1 |
| **G** | **I** | Family members are detained | 1 |
| **G** | **II** | H maintains contact with K | 1 |
| **H** | **I** | X trusts A and E most after arrest | 1 |
| **H** | **II** | B distrusts others | 1 |
| **H** | **III** | E maintains contact with X | 1 |
| **H** | **IV** | B maintains contact with X | 1 |
| **H** | **V** | E trusts A and X | 1 |
| **H** | **VI** | A is detained | 1 |
| **H** | **VII** | B is detained | 1 |
| **I** | **I** | A contacts the kingpin | 1 |
| **I** | **II** | A trusts B more than others | 1 |
| **J** | **I** | A contacts the kingpin | 1 |
| **J** | **II** | Trust between A and B increases | 1 |
| **J** | **III** | A is detained | 1 |
| **J** | **IV** | B is detained | 1 |

**Key Outputs of Phase II**

**Conceptual Model:** A theoretical representation of kingpin replacement dynamics in criminal networks. This model outlines agent roles, behavioral dependencies, and key mechanisms such as trust adaptation and leadership emergence.
**Validated Conceptual Model:** An updated conceptual model refined through qualitative validation against real-world case files (A, B, C). This version captures observed patterns of trust reorganization, leadership succession, and the structural resilience of criminal networks.
**Calibrated ABM:** An agent-based model calibrated using empirical data and calibration statements from case files A, B, and C. This model uses optimized parameters (e.g., trust dynamics, kingpin thresholds) and reflects observed criminal network adaptations during leadership transitions. See Table 14 for a guide on how the case files were distributed within the toy model.

## Phase III – Validation

The Validation Statements, validation thresholds and calibrated ABM in Phase II then lead into the Validation in Phase III, which is the process leading to the validated ABM. The validation process follows three key steps: Initialization, Simulation Execution (run the ABM through all four phases (stable, intervention, who-done-it, cooldown) with calibrated parameters.) and Outcome Comparison (evaluate simulated outcomes against VSs (e.g., leader emergence, network restructuring).)

Each VS is operationalised as a binary predicate (satisfied / not satisfied). Model performance is quantified as the proportion of VSs satisfied across repeated simulation runs (n = 48), reflecting stochastic variability in the underlying ABM. Consistent with established practice in agent-based model validation, we define a target performance threshold in the range of 80–90% satisfaction; for this proof-of-concept application, a working benchmark of 85% is adopted.

Given the limited and fully qualitative nature of the dataset, this study should be interpreted as an initial demonstration of the FREIDA validation procedure rather than a fully data-constrained empirical validation. Under these conditions, the model satisfies 3 out of 5 VSs (60%). This outcome reflects the restricted scope and heterogeneity of the available case material rather than a failure of the modelling approach.

The model reproduces several higher-level structural dynamics, including leadership succession following kingpin removal and associated network reorganisation and trust reconfiguration. However, it does not consistently reproduce more granular or context-specific validation statements, which appear to depend on locally variable or under-specified mechanisms in the underlying qualitative evidence.

Within the FREIDA framework, validation is not treated as a binary pass/fail criterion but as a diagnostic assessment of behavioural fidelity. Partial VS satisfaction is expected in settings characterised by sparse, heterogeneous expert knowledge and structurally uncertain mechanisms. The validation results therefore serve to identify robust emergent patterns that are consistently reproduced by the model, as well as components of the system that remain under-constrained and require further empirical refinement.

Overall, the results demonstrate that the proposed pipeline can translate qualitative expert knowledge into a computational ABM that reproduces key emergent system behaviours, while explicitly exposing the limits of the available evidence and the resulting uncertainty in model generalisation.

**Validated Model (Final Output)**

The validated model is an agent-based simulation capable of reproducing observed kingpin replacement dynamics across multiple real-world scenarios. It accurately reflects both short-term behavioral shifts (e.g., violence capital increases) and long-term structural changes (e.g., leadership consolidation). A refined set of model parameters optimized through stochastic search (SPSA), with a focus on trust dynamics and kingpin selection mechanisms. Key adjustments include the calibration of the family-tie trust parameter ($\varphi$) and the minimum kingpin attribute threshold ($\varkappa$).

The key outputs in Phase III are the validated model, the quantitative validation report, and an updated parameter set.

**Validation Statements**

Table 12.a. presents the VSs distilled from expert knowledge for the toy model. These statements define expected model behaviors and outcomes at a broader system level, focusing on network-wide properties and trends. They specify criteria related to trust dynamics and network connectivity across different case scenarios. This table is essential for the Model Validation (Phase III) section, where the model's ability to generalize and replicate observed patterns is assessed. Table 12.b. presents the additional VSs from case files E-J that were not used in the validation of the toy model.

**Table 12.a.:** Validation Statements of case files A-D distilled from the expert knowledge for the toy model. Each VS represents a (partial) score point.

| Case | ID | Validation Statements | Maximum score |
|---|---|---|---|
| **A** | **V** | All high trust values (>0.8) should have increased or at least remained the same directly after the killing | 1 |
| **A** | **VII** | Average trust among the orphans increased after 1 year (VS) | 1 |
| **B** | **VIII** | All high trust values (>0.8) should have increased or at least remained the same directly after the killing | 1 |
| **B** | **III** | The average trust among the orphans increases after the liquidation (measured at 364 days after) | 1 |
| **D** | **I** | Y is new murderbroker by the end of the simulation | 1 |
| **C** | **V** | The trust between the orphans increases by 15% before the new replacement is chosen | 1 |
| **C** | **IV** | All high trust values (>0.8) should have increased or at least remained the same directly after the killing | 1 |
| **D** | **V** | All high trust values (>0.8) should have increased or at least remained the same directly after the killing (partial score possible) | 1 |
| **A** | **VIII** | Connectivity and/or trust values among the non-kingpin nodes changed significantly (at least 0.1) | 1 |

**Table 12.b.:** Validation Statements of case files E-J distilled from the expert knowledge for validation purposes. Each VS represents a (partial) score point.

| Case | ID | Validation Statements | Maximum Score |
|---|---|---|---|
| **E** | **IX** | D1 becomes structurally central | 1 |
| **E** | **X** | Network degree distribution behaves as expected | 1 |
| **F** | **XI** | Network clustering behaves as expected | 1 |
| **F** | **XII** | Tie stability is preserved | 1 |
| **G** | **III** | G retains customers after disruption | 1 |
| **G** | **IV** | G becomes kingpin | 1 |
| **G** | **V** | I never becomes kingpin | 1 |
| **G** | **VI** | Degree structure is preserved | 1 |
| **G** | **VII** | Clustering is preserved | 1 |
| **G** | **VIII** | Tie stability is preserved | 1 |
| **H** | **VIII** | A becomes kingpin candidate | 1 |
| **H** | **IX** | C–E–F trio forms | 1 |
| **H** | **X** | Trust increases within subgroup | 1 |
| **H** | **XI** | Degree distribution behaves as expected | 1 |
| **H** | **XII** | Clustering behaves as expected | 1 |
| **I** | **III** | A becomes kingpin candidate | 1 |
| **I** | **IV** | Network reorganizes after one month | 1 |
| **I** | **V** | A becomes new kingpin | 1 |
| **I** | **VI** | Degree distribution behaves as expected | 1 |
| **I** | **VII** | Clustering behaves as expected | 1 |
| **I** | **VIII** | Tie stability is preserved | 1 |
| **J** | **V** | A becomes kingpin candidate | 1 |
| **J** | **VI** | A becomes new kingpin | 1 |
| **J** | **VII** | Former kingpin remains influential | 1 |
| **J** | **VIII** | Degree distribution behaves as expected | 1 |
| **J** | **IX** | Clustering behaves as expected | 1 |
| **J** | **X** | Tie stability is preserved | 1 |

## Post-Phase III – Sensitivity Analysis (SA) and Uncertainty Quantification (UQ)

The last steps analyze the model's robustness and assesses the impact of uncertainty on outcomes. If the validated model at the end of phase III is satisfying, then the SA and UQ are not needed, and the validated ABM can be optionally enriched by adding more dynamics in the ODD+D. If the model does not meet validation criteria, then SA and UQ is needed before it can be enriched further.

These parameters are pivotal for modeling the network's functionality and resilience. For instance, customs officers and gatekeepers are crucial in ensuring the undetected passage of cocaine, while coordinators and financers play essential roles in organizing and funding the operations. By quantifying these roles, Table 16 offers a comprehensive view of the criminal hierarchy and its operational mechanics. This structured approach aids in simulating various scenarios, such as the impact of the removal of key figures, and helps in understanding the potential shifts and adaptations within the network.

### Sensitivity Analysis and Uncertainty Quantification

Sensitivity Analysis (SA) and Uncertainty Quantification (UQ) in the toy model were conducted using a One-At-a-Time (OAT/OFAT) approach. This choice reflects the intentionally minimal and illustrative nature of the toy model, rather than the full methodological ambition of the FREIDA framework. In the full FREIDA pipeline, SA and UQ are envisioned as more comprehensive procedures (e.g., global sensitivity analysis and sampling-based uncertainty propagation) that account for parameter interactions and high-dimensional effects.

Here, however, OFAT was used for computational simplicity and interpretability, allowing us to isolate the influence of individual parameters, such as minimum kingpin attributes (Φ) and trust-related factors, on model outcomes. These parameters were identified as particularly relevant in the qualitative case material and in the construction of ESBs.

While OFAT does not capture higher-order parameter interactions, it is sufficient at this stage to demonstrate directional sensitivities and to support iterative model refinement. In this sense, the analysis is intended as a diagnostic tool for the toy model rather than a full uncertainty quantification exercise.

For UQ, a forward approach sampled multiple parameter sets, revealing how input combinations influenced model outputs. Results showed Φ and trust-related parameters as the most sensitive, underscoring the need to refine trust dynamics, especially family-tie trust updates after contagion related to loyalty and coercion. Both SA and UQ highlighted the critical role of accurately modeling trust in criminal networks.

The global SA reveals that the minimum attributes for kingpin ($\varkappa$) parameter has the highest influence on model outcomes, with a global sensitivity score of approximately 0.145. This is significantly greater than other parameters, such as minimum trust contagion, minimum traits suggestion, Tau, Psi, Phi, and T, each of which exhibit values around 0.015, indicating comparatively minor impacts.

Local SA shows a different pattern. The minimum trust contagion parameter is the most sensitive, with a local sensitivity score of 0.006, underscoring its critical role in scenarios involving trust breakdowns. Phi, which governs family-tie trust updates, follows with a score of 0.0045, highlighting its importance in post-replacement dynamics especially when coercion or violence overrides relational trust. Other parameters such as Tau, Psi, minimum kingpin attributes ($\varkappa$), and T range from approximately 0.002 to 0.003, indicating more moderate effects.

These findings suggest that both $\varkappa$ and trust-related dynamics, particularly minimum trust contagion and Phi, are essential to model fidelity, especially in high-stakes scenarios involving violence or coercion.

The unsatisfactory validation score further supports the need for refinement. Incorporating the results from both global and local SA, we identify $\varkappa$ and trust-related parameters as high-priority targets for improvement. FREIDA's iterative framework will be leveraged to guide this process.

At later possible extensions, Phase I will focus on redefining $\varkappa$ and trust-related parameters to better reflect their impact. Phase II should collect more detailed data on kingpin dynamics, including behaviors driven by trust and coercion. Phase III will recalibrate the model, improving its responsiveness to the refined parameters. Finally, Post-Phase III will validate the updated model using new data, ensuring enhanced accuracy in simulating criminal network dynamics.

Uncertainty analysis reveals that the model is most sensitive to trust-related parameters ($\psi$ and $\varphi$), reflecting the central role of interpersonal relationships in kingpin replacement. Variability in model outcomes suggests that small changes in trust dynamics post-intervention lead to significantly different leadership structures. Policymakers should interpret predictions with caution, especially in cases involving coercion or fragmented trust.

**Start next iteration**

In the current execution of the model, no formal SA was performed, which would involve refining the model by removing and changing parameters until it functions with the minimum necessary parameters while maintaining the same outcomes. Formal SA would likely remove repetitive parameters from Calibration and Validation Statements, such as calibration statements BIII and BVIII, which similarly test trust increase after kingpin removal. Adjusting parameters XIV and XV for kingpin or murderbroker candidates (Appendix II, Table 9) is not expected to change outcomes since the final parameters (XVI and XVII) determine candidate fitness. Candidates meeting the first but not the second set cannot become the final kingpin or murderbroker. Edmonds argues that formal model understanding should not overshadow model adequacy, with SA being essential for assessing reliability. Using a partially understood model with SA is preferable to not modeling or using unreliable models.

## Additional Contextual Information

### Case file analysis

When distilling behavioral rules from case files, clearly stating cause-and-effect relationships is crucial for modelers. Instead of simply noting "Agent X did A, then Agent Y did B," specify causation ("Y did B *because* X did A"), independence ("Y did B *independently*"), or acknowledge uncertainty. This explicitness avoids modeller bias or increased model uncertainty during the implementation of if-then-else behavioral rules. Additionally, annotating cases with temporal scales (time between actions, overall timeframe, identifiable phases) and spatial scales (network size, geographical extent) is vital for aligning with the ODD+D and for accurately modeling time delays and spatial dynamics within the simulation.

The most relevant components of case files include the identification of a network topology (scope and scale of a network) and the agent and group specific behaviours (translation into behavioural rules). In Table 13, the case files are broken down into even more specific categories and examples as well as concrete details provided.

Table 13 provides examples of how information from case files is translated into key concepts for agent-based modeling. It details how textual descriptions are converted into structured data related to time, agent characteristics, behavior, rules, network topology, and inter-agent ties. This process is crucial for bridging the gap between qualitative data and computational model requirements. This table supports the Model Development (Phase II), specifically the process of extracting and formalizing information for model implementation.

**Table 13:** Examples of case file translation

| Concept | Details | Example | Concretely |
|---|---|---|---|
| **Time** | Must include timeline, scales, jumps, end time | First week, first months, after 1 year | Months, days, etc |
| **Agents** | Roles, specific, descriptions | Orphans, amount of agents, etc | Specific key agents, successors, potential replacements |
| **Behaviour** | Agent behaviour (motivations) | Specific roles (social and business) define the agent behaviour | Roles within the network and responsibilities (orphans, three categories of roles, etc.) |
| **Rules** | General agent and network rules | Events happen at set timesteps, agents switch from one to other behaviour patterns using triggers | Orphans choose the new successor based on the selection-rules as determined by the case file |
| **Topology** | Network growth, demography of agents, etc. | Connections are added based on triggering moments | When kingpin is removed, ties change on the basis of the trust. When a new kingpin is chose, every node automatically establishes a tie with the new kingpin based on their role |
| **Ties** | Tie description, changes, etc | Severed ties, tie connection, social and business layer, etc. | Agent ties depending on roles, trust, financial-, criminal-capital, violence, orphan connections to new replacement |

**Example case file**

Below is an example case file in the style of case files A-J. The below case file has been synthetically generated using natural language processing techniques. It has not been used in the toy model. For a breakdown on how the case files were used in the toy model, please consult Table 14.

1. Context network:

1.1. Socio-cultural scene: Moroccan and Dutch

1.2. Geo scene: the Netherlands and Colombia. The network has its roots in the New West in Amsterdam.

1.3. Criminal markets: cocaine, heroine and money laundering

1.4. Network structure: > 100 members.

1.5. Violence exposure: mild violence exposure. The network is involved in some conflicts.

2. Description X: X is of Moroccan origin and is the kingpin of a contingent of assassins. Part of those assassins comes from a Dutch background. Also X started as a killer, but he improved his skills as a network organizer. He has a particularly high IQ.
3. T0 Intervention: X is killed.
4. T1 month after intervention

Behaviour orphan A: X is replaced by A. They had a relationship of mutual respect based on criminal trust. X trained A to be aggressive and without limits and to use violence only when necessary. A takes the place of X because he is the best fusion of organization and violence.

Behaviour orphan B: B is a direct contact of A and trusted person. After X killing, B in couple with A is busy in reorganizing the network. B is the main man of violence in the network.

Behaviour orphan C: C another assassin that grew up in the New West. He is the main suspect in the network of having killed X.

Behaviour orphan D: D manages cocaine imports from Colombia and exports towards european countries.

Behaviour orphan E: E is an experienced assassin, considered the right arm of X.

Behaviour orphan F: F works in strict contact with E when X is killed.

5. T2 (four months after killing)

Behaviour orphan A: he has the reputation of a reliable organizer. He is able to direct the network and being portrayed in the gangster rap scene. Although he travels a lot he goes very often to his old neighborhood, which is New West. He carries out violent jobs, even if when these jobs are risky.

Behaviour orphan B: B is in contact with A and organizes violent jobs.

Behaviour orphan C: C communicates with A and B and implements violent jobs.

Behaviour orphan D: to D is assigned the management of the cocaine imports from Colombia.

Behaviour orphan E: E works together with B and D.

Behaviour orphan F: F works in strict contact with the members of A’s network.

6. T3 (years after killing)
Behaviour orphan A: he carries out an assignment, but is arrested. In restraints he cannot communicate with the rest of the network apart from B, through his lawyers. For both X and A is almost impossible to communicate with the rest of the world. This happens when a criminal is posed in EBI.

Behaviour orphan B: B communicates with A about X.
Behaviour orphan C: no communication with A.
Behaviour orphan D: no communication with A.
Behaviour orphan E: no communication with A.
Behaviour orphan F: no communication with A.

To clarify the use of the case files in the toy model, please refer to Table 14:

**Table 14:** Case File User Guide

| Case Files | Size | Role in the Toy Model |
|---|---|---|
| **A, B, C** | 3 Files | Calibration Set: Used to actively tune the model parameters. |
| **D** | 1 File | Validation Set: Saved to test out-of-sample performance within the core group. |
| **A-D** | 4 Files | Core Pool |
| **E-J** | 6 Files | External Hold-Out Set: Used exclusively for secondary robustness checks and cross-case generalization. |
| **A-J** | 10 Files | The Full Empirical Dataset: The entire scope of qualitative case data processed by the framework. |

### Value Network

When the VN is initialized, each role is added with their probability of replacement, connectivity, and intrinsic capital (criminal, violence, financial). Connectivity refers to the probability of forming random other connections outside their value chain.

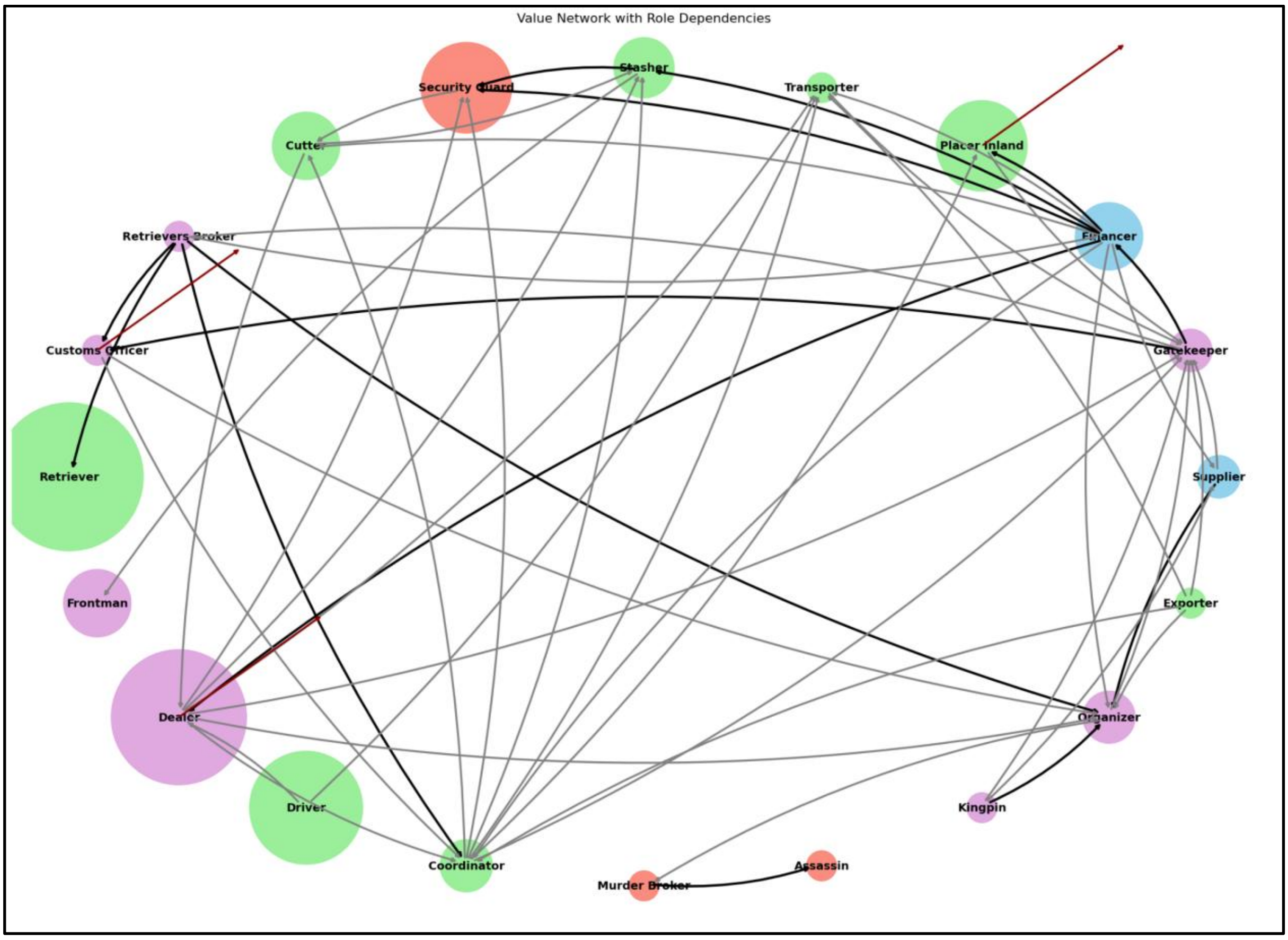

**Figure 7:** The Value Network of the different roles and their dependencies. The size of the nodes indicates the relative frequency of each role occurring in the toy model. Thick edges indicate a mutual dependency, thin edges are directional (please refer to the dependencies between roles for specifications) and the short edges to the right of nodes that do not connect to another node indicate a self-edge, thus that role is dependent on knowing others with the same role. Node colors distinguish the type of activity associated with each role: financial roles, such as the Financer and Supplier, are blue; logistical roles, including Transporter, Retriever, and Stasher, are green; roles with high criminal capital, like the Kingpin, Organizer, and Gatekeeper are purple; and violent roles, such as Assassin and Security Guard, are orange. Node size reflects the relative number of individuals performing each role. For example, roles like Dealer and Retriever, which involve a large number of people, are represented with significantly larger nodes, while highly centralized or strategic roles like Kingpin or Exporter are smaller.

## Model Narrative

The model narrative is depicted in Table 15. We begin with the initialization phase according to the specifications as set by the modellers and domain experts. The four stages (initialization, removal of the kingpin, searching stage, and finally instating of a new kingpin) follow in succession, and return the network back into the stable state it began with.

Table 15 provides a concise overview of the toy model's phases, delineating both the conceptual and computational stages. It describes the sequence of events within the model, from the initial setup to the intervention and subsequent network adjustments, and specifies the timing of each phase. This table helps to clarify the temporal dynamics of the model and provides a roadmap for understanding the simulation's progression, supporting the explanation of the model's design and functionality in the Model Development (Phase II).

**Table 15:** A brief overview of each phase of the toy model, with a short description as well as an overview of the time of commencement for the respective phase

| Conceptual stage (model phases) | Computational stage | Description | Time scale |
|---|---|---|---|
| Stable stage | Initialization | All agents and ties are being initialised into the model according to their initialization specifications. | Step 0 |
| Intervention | Removal of the kingpin | The initial kingpin and his ties are removed. This step marks the dynamic beginning of the simulation. | Step 0 |
| Who-done-it | Searching phase | The orphans evaluate the potential replacements until a suitable replacement is found. | Commences between 10 and 30 steps after kingpin removal. |
| Cooldown | Instating of a new kingpin | The new kingpin officially gets picked and changes roles from their old role to the new role. | Commences when a new kingpin is picked. |
| Cooldown/ Stable stage | Model updates | Updates according to the new change in the network. | Commences after the new kingpin is picked |

The created model contains the cases of four provided case files (A-D), and accurately describes the events of the agents within the case files throughout one year. Specifically, the removal of the kingpin or murderbroker commences at step 0, the events within step 1 (up to 31 days after removal) are followed closely, and the model is once again assessed after 365 days (from step 0). Per time step, the model undergoes the changes in tie-connectivity and orphan-behaviour. Case files E-J were not included in the creation of the toy model, other than using the calibration statements of cases E-J for the calibration trajectories (see Figure 4.a.).

# Appendix II: Parameters and Outcomes

In the analysis of criminal networks involved in cocaine trafficking, understanding the roles and interactions within these networks is crucial. To systematically capture and evaluate these dynamics, structured interviews with domain experts were employed. This approach is exemplified in Table 16, which provides an overview of the roles identified in the toy model, along with their respective parameters for criminal, violence, and financial capital. Each role, from customs officers to kingpins, is assigned specific values that reflect their importance and influence within the network.

We follow the phases of FREIDA and explain additional outputs and processes specific to the toy model.

Table 16 presents an overview of the parameters extracted per agent role through structured interviews with domain experts, focusing on Criminal, Violence, and Financial Capital. These capital values, expressed qualitatively as low, medium, or high, characterize the attributes of different roles within the criminal network. This table provides crucial input for defining agent characteristics and behaviors within the toy model, supporting the Model Development (Phase II) section by informing the operationalization of agents.

**Table 16:** An overview of the parameters extracted per role through structured interviews with domain experts used in the toy model. Capitals are expressed in low (0-0.3), medium (0.3-0.7) and high (0.7-1).

| **Role** | **Criminal Capital** | **Violence Capital** | **Financial Capital** | **Description** |
|---|---|---|---|---|
| (Corrupt) customs officer | medium | medium | low | Ensures that the cocaine is not detected when entering the import country |
| Gatekeeper | high | low | medium | Decides what and who gets through certain gates at (air)ports |
| Transporter | low | medium | medium | Transports the cocaine from the country of origin to the import country |
| Distributer | medium | medium | medium | Person distributing the cocaine through the network |
| Coordinator | high | medium | medium-high | Coordinates the transport within the country of origin and the country of import |
| Exporter | medium | medium | medium | Exports the cocaine from the country of origin (usually in South America) |
| Financer | high | medium | high | Finances cocaine operations |
| Kingpin | high | medium | medium | Most authorative and important person in the criminal network, with a high criminal capital |
| Producer | high | medium | medium | Produces cocaine |
| Organizer | high | medium | high | Organizing operations within the cocaine network |
| Broker of Retrievers | high | medium | medium | Knows and hires cocaine retrievers |
| Broker | high | medium | medium | Knowledgeable about agents with needed roles and able to connect roles to each other |
| Cutter | low | low | low | Cuts cocaine and mixes it with other substances to increase profits or change the drugs effect |
| Driver | low | low | low | Transports the cocaine to or from the (air)ports |
| Placer Inland | low | medium | low | Coordinates the amount of cocaine to be brought to each place within the import country |
| Stasher | low | medium | low | Stores the cocaine until it is ready to be sold |
| Frontman | low | medium | low | Represents the criminal organization and tries to make its activities seem acceptable to the public |
| Retriever | low | medium | low | Often minors that take out the drugs from containers for criminal organizations |
| Murderbroker | low-medium | high | low-medium | Person organizing and hiring assassins |
| Assassin | low-medium | high | low-medium | Person liquidating other agents |
| Dealer | medium | medium | medium | Person selling cocaine to end-customers |